\documentclass[10pt,twocolumn,letterpaper]{article}

\usepackage[pagenumbers]{cvpr} 

\usepackage[dvipsnames,svgnames,x11names]{xcolor}
\usepackage[dvipsnames]{xcolor}

\usepackage{graphicx}
\usepackage{amsmath}
\usepackage{amssymb}
\usepackage{booktabs}
\usepackage{multirow} 
\usepackage{multicol}
\usepackage{wrapfig}
\usepackage{etoc}
\usepackage{minitoc}
\usepackage{bm}
\usepackage[accsupp]{axessibility}

\definecolor{cvprblue}{rgb}{0.21,0.49,0.74}
\usepackage{xcolor}
\PassOptionsToPackage{hyphens}{url}\usepackage[pagebackref=true,breaklinks=true,letterpaper=true,colorlinks,
  citecolor=cvprblue,bookmarks=false]{hyperref}

\usepackage[capitalize]{cleveref}
\crefname{section}{Sec.}{Secs.}
\Crefname{section}{Section}{Sections}
\Crefname{table}{Table}{Tables}
\crefname{table}{Tab.}{Tabs.}

\def\themodel{ConsistDreamer\xspace} 

\makeatletter
\renewcommand\paragraph{\@startsection{paragraph}{4}{0mm}
                                   {0ex}
                                   {-1em}
                                   {\normalfont\normalsize\bfseries}}
\makeatother

\def\sv{{\small\color{DarkOrange}{\bf SV}}\xspace}
\newcommand{\svat}[1]{\sv}

\def\paperID{6107} 
\def\confName{CVPR}
\def\confYear{2024}

\title{\themodel: 3D-Consistent 2D Diffusion for High-Fidelity Scene Editing}

\author{Jun-Kun Chen$^{1}$$^\dagger$ \qquad Samuel Rota Bulò$^{2}$ \qquad Norman Müller$^{2}$ \qquad Lorenzo Porzi$^{2}$ \vspace{0.1em} \\ Peter Kontschieder$^{2}$ \qquad Yu-Xiong Wang$^{1}$ \vspace{0.1em} \\ 
    $^1$University of Illinois Urbana-Champaign \qquad $^2$Meta \vspace{0.1em}\\
    {\tt \hspace{0mm}\{junkun3,yxw\}@illinois.edu} \vspace{0.1em}\\ {\tt \{rotabulo,normanm,porzi,pkontschieder\}@meta.com}
}

\begin{document}

\twocolumn[{
\maketitle
\vspace{-5.0mm}
\renewcommand\twocolumn[1][]{#1}
    \centering
    
    \includegraphics[width=1.0\linewidth]{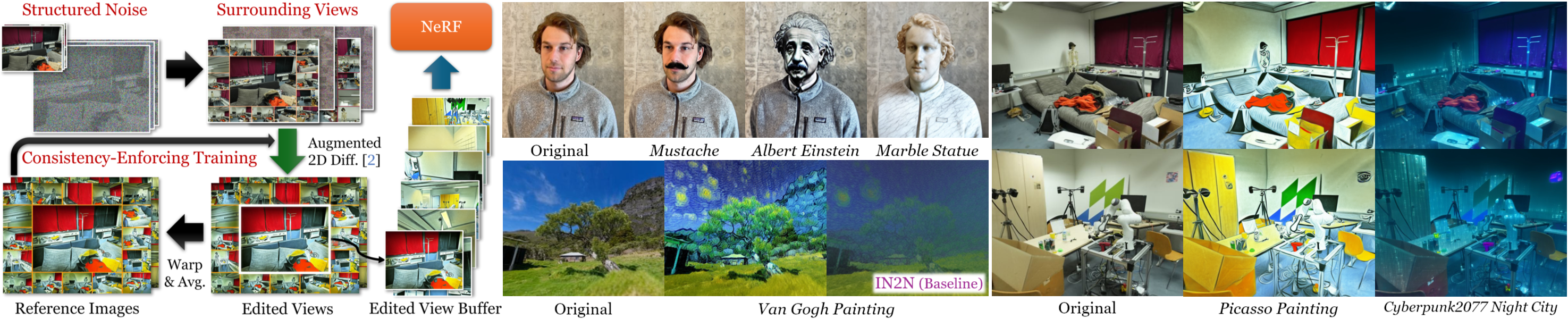}
    \captionof{figure}{\textbf{Our {\themodel} lifts 2D diffusion with 3D awareness and consistency}, achieving high-fidelity instruction-guided scene editing with superior sharpness and detailed textures. \textbf{Left:} The \textcolor[RGB]{192,0,0}{three synergistic components} within \themodel that enable 3D consistency. \textbf{Right:} State-of-the-art performance of \themodel across various editing tasks and scenes, especially when \textcolor[RGB]{160,47,147}{prior work (\eg, IN2N~\cite{in2n})} fails and in challenging large-scale indoor scenes from ScanNet++~\cite{scannetpp}. More results are on our \href{https://immortalco.github.io/ConsistDreamer/}{project page}.} 
\label{fig:teaser}
    \vspace{2mm}
}]

\begin{abstract}
\vspace{-2mm}
This paper proposes \themodel{} -- a novel framework that lifts 2D diffusion models with 3D awareness and 3D consistency, thus enabling high-fidelity instruction-guided scene editing. To overcome the fundamental limitation of missing 3D consistency in 2D diffusion models, our key insight is to introduce three synergistic strategies that augment the input of the 2D diffusion model to become 3D-aware and to explicitly enforce 3D consistency during the training process. Specifically, we design surrounding views as context-rich input for the 2D diffusion model, and generate 3D-consistent structured noise instead of image-independent noise. Moreover, we introduce self-supervised consistency-enforcing training within the per-scene editing procedure. Extensive evaluation shows that our \themodel achieves state-of-the-art performance for instruction-guided scene editing across various scenes and editing instructions, particularly in complicated large-scale indoor scenes from ScanNet++, with significantly improved sharpness and fine-grained textures. Notably, \themodel stands as the first work capable of successfully editing complex (e.g., plaid/checkered) patterns. Our project page is at \href{https://immortalco.github.io/ConsistDreamer/}{\textit{immortalco.github.io/ConsistDreamer}}.
\end{abstract}
\vspace{-3mm}
\renewcommand{\thefootnote}{\fnsymbol{footnote}}
\footnotetext[2]{Work started during an internship at Meta Reality Labs Zurich.}
\renewcommand{\thefootnote}{\arabic{footnote}}
\section{Introduction}
\label{sec:intro}

With the emergence of instruction-guided 2D generative models as in~\cite{ip2p}, it has never been easier to generate or edit images. Extending this success to 3D, \ie, instruction-guided 3D scene editing, becomes highly desirable for artists, designers, and the movie and game industries. Nevertheless, editing 3D scenes or objects is inherently challenging. The absence of large-scale, general 3D datasets makes it difficult to create a counterpart generative model similar to~\cite{ip2p} that can support arbitrary 3D scenes. Therefore, state-of-the-art solutions~\cite{dreamfusion,in2n} circumvent this challenge by resorting to generalizable 2D diffusion models. This approach, known as \emph{2D diffusion distillation}, renders the scene into multi-view images, applies an instruction-conditioned diffusion model in 2D, and then distills the editing signal back to 3D, such as through a neural radiance field (NeRF)~\cite{in2n,vica,csd}.

However, a fundamental limitation of this solution is \emph{the lack of 3D consistency}: a 2D diffusion model, acting independently across views, is likely to produce inconsistent edits, both in color and shape.
For example, a person in one view might be edited to be wearing a red shirt, while appearing in a green shirt in another view. 
Using these images to train a NeRF can still produce reasonable edits, but the model will naturally converge towards an ``averaged'' representation of the inconsistent 2D supervision, and lose most of its details and sharpness.
A commonly observed failure mode is that of regular (\eg, checkered) patterns, which completely disappear once distilled to 3D due to misalignments across views.
Generating consistent multi-view images thus becomes crucial for achieving high-fidelity 3D scene editing. 

While largely overlooked in prior work, our investigation reveals that the source of inconsistency is \emph{multi-faceted}, and  primarily originates from the \emph{input}. \textbf{(1)} As the 2D diffusion model can only observe a single view at a time, it \emph{lacks sufficient context} to understand the entire scene and apply consistent editing. \textbf{(2)} The editing process for each image starts from \emph{independently generated Gaussian noise}, which brings challenges to consistent image generation. Intuitively, it is difficult to generate consistent multi-view images by denoising inconsistent noise, and even for a single view, it may not always yield the same edited result. \textbf{(3)} The input to the 2D diffusion model contains \emph{no 3D information}, making it much harder for the model to reason about 3D geometry and to share information across different views of the scene, even when made available to it.

Motivated by these observations, we propose \themodel{} -- a novel framework to achieve 3D consistency in 2D diffusion distillation. \themodel introduces three \emph{synergistic} strategies that augment the input of the 2D diffusion model to be 3D-aware and enforce 3D consistency in a self-supervised manner during the training process. 

To address the limited context issue within a single view, our framework involves incorporating context from other views. We capitalize on the observation that 2D diffusion models inherently support ``composed images,'' where multiple sub-images are tiled to form a larger image. Given the capability of the self-attention modules in the UNet of the 2D diffusion model to establish connections between the same objects across different sub-images, each image can be edited with the context derived from other images. Therefore, we leverage the composed images to construct a {\em surrounding view} (Fig.~\ref{fig:teaser}), where one large, central main view is surrounded by several small reference views. This approach allows us to edit the main view with the context from reference views, and vice versa. Doing so not only enriches the context of the scene in the input, but also enables the simultaneous editing of multiple views.

Regarding the noise, we introduce {\em 3D-consistent structured noise} (Fig.~\ref{fig:teaser}), with the key insight of generating consistent noise for each view once at the beginning. Specifically, we generate and fix Gaussian noise on the surface of the scene objects, and then render each view to obtain the 2D noise used for the image at that view in all subsequent diffusion generations. This approach aligns with existing 3D diffusion work~\cite{diffrf} which also generates noise in 3D at the beginning of a generation. Ensuring that the denoising procedure starts with consistent noise substantially facilitates the process of achieving consistent images by the end.

The combination of surrounding views and structured noise provides the 2D diffusion model with 3D consistent input, yet it is insufficient. An explicit enforcement of 3D consistency is also required during the learning process. To this end, we propose {\em self-supervised consistency-enforcing training} within the per-scene editing procedure (Fig.~\ref{fig:teaser}). We augment the 2D diffusion model by a ControlNet~\cite{controlnet} that introduces 3D positional embedding to make it 3D-aware. Inspired by~\cite{consistnet,ivid}, we perform warping and averaging for all sub-views in the edited surrounding view image. This process yields a surrounding view of 3D consistent sub-views used as the self-supervision target. To further achieve ``cross-batch consistency'' -- consistency between different batches in different generations -- we perform multiple generations in parallel, and construct consistent target images from all sub-views in all generated surrounding view images, so as to supervise all generations collectively. After consistency-enforcing training, the 2D diffusion model is able to generate consistent multi-view images. Consequently, a trained NeRF will not have to smooth out inconsistencies, but ultimately converge to sharp results preserving fine-grained details.

Empowered by such a 3D-consistent 2D diffusion model, our \themodel achieves high-fidelity and diverse instruction-guided 3D scene editing {\em without} any mesh exportation and refinements or a better scene representation like Gaussian Splatting \cite{3dgs}, as shown in Fig.~\ref{fig:teaser}. Compared with previous work, the editing results of \themodel exhibit significantly improved sharpness and detail, while preserving the diversity in the original 2D diffusion model's~\cite{ip2p} editing results. Notably, \themodel stands as the \emph{first} work capable of successfully editing complex (\eg, checkered) patterns. Moreover, \themodel demonstrates superior performance in complicated, high-resolution ScanNet++~\cite{scannetpp} scenes -- an accomplishment where state-of-the-art methods faced challenges in achieving satisfactory edits.

\textbf{Our contributions} are three-fold. \textbf{(1)} We introduce \themodel, a simple yet effective framework that enables 3D-consistent instruction-guided scene editing based on distillation from 2D diffusion models. \textbf{(2)} We propose three novel, synergistic components -- structured noise, surrounding views, and consistency-enforcing training -- that lift 2D diffusion models to generate 3D-consistent images across all generated batches. Notably, our work is the \emph{first} that explores cross-batch consistency and denoising consistency in 2D diffusion distillation and attains these through manipulating noise. \textbf{(3)} We evaluate a range of scenes and editing instructions, achieving state-of-the-art performance in both, scenes considered by previous work and more complicated, large-scale indoor scenes from ScanNet++.
\section{Related Work}
\label{sec:related}

\paragraph{NeRF-Based Scene Editing.} Neural radiance field (NeRF)~\cite{nerf} and its variants~\cite{ibrnet,pixelnerf,pointnerf,mvsnerf,mipnerf,nerfren,refnerf,neuralsparse} are widely-used approaches to representing scenes. NeRF leverages neural networks or other learnable architectures to learn to reconstruct the 3D geometry of a scene only from multi-view images and their camera parameters, and support novel view synthesis. With the development of NeRF, editing a NeRF-represented scene is also deeply studied, covering different types of editing objectives and editing operation indicators, \emph{a.k.a.}, ``user interfaces.'' Some methods~\cite{editnerf,objectnerf,distillnerf} support editing the position, color, and/or shape of a specific object indicated by users through a pixel, a text description, or a segment, \etc Another line of work~\cite{deformingnerf,cagenerf,nerfediting,neuraleditor} studies human-guided shape editing, which allows users to indicate a shape editing operation with a cage or point cloud provided by the model. The task we investigate is instruction-guided scene editing, which allows users to indicate the editing operation through instructions in natural language. The first work in this direction is NeRF-Art~\cite{nerfart}, which mainly focuses on style transfer instructions, and uses pre-trained CLIP~\cite{clip} as the stylization loss for the instruction-indicated style. More recent work~\cite{in2n,i3d23d,csd,editdiffnerf,dreameditor} leverages diffusion models~\cite{diffusion} instead of CLIP to benefit from powerful diffusion models and support more general instructions.

\paragraph{Distillation-Based 3D Scene Generation.} Lacking 3D datasets to train powerful 3D diffusion models, current solutions distill the generation signal from a 2D diffusion model to exploit its ability in 3D generation. DreamFusion~\cite{dreamfusion} is the first work in this direction, which proposes score distillation sampling (SDS) to distill the gradient update direction (``score'') from 2D diffusion models, and supports instruction-guided scene generation by distilling a pre-trained diffusion model~\cite{sd}. HiFA~\cite{hifa} proposes an annealing technique and rephrases the distillation formula to improve the generation result. Magic3D~\cite{magic3d} improves the generation results by introducing a coarse-to-fine strategy and a mesh exportation and refinement method. ProlificDreamer~\cite{prolificdreamer} further improves the generation results by introducing an improved version of SDS, namely variational score distillation (VSD), to augment and fine-tune a pre-trained diffusion model and use it for generation. 

\paragraph{Diffusion Distillation-Based 3D Scene Editing.} Similar to \cite{sd} for instruction-guided generation tasks, another diffusion model \cite{ip2p} was proposed for instruction-guided image editing, by generating the edited image conditioned on both the original image and the instruction, which is therefore compatible with SDS~\cite{dreamfusion}. Instruction 3D-to-3D~\cite{i3d23d} uses SDS with \cite{ip2p} to support instruction-guided style transfer on 3D scenes. Instruct-NeRF2NeRF (IN2N)~\cite{in2n} adopts another way to operate the 2D diffusion model, similar to the rephrased version of SDS in HiFA~\cite{hifa}, which iteratively generates edited images to update the NeRF dataset for NeRF fitting, and supports more general editing instructions such as object-specific editing. ViCA-NeRF~\cite{vica} proposes a different pipeline to first edit key views and then blend key views and apply refinement. Edit-DiffNeRF~\cite{editdiffnerf} augments the diffusion model and fine-tunes it with a CLIP loss to improve the success rate of editing. DreamEditor~\cite{dreameditor} utilizes a fine-tuned variant of \cite{dreambooth} instead of \cite{ip2p} and focuses on object-specific editing. 

\paragraph{Consistency in Distillation-Based Pipelines.} When distilling from 2D diffusion to perform 3D generation or editing, 3D awareness and 3D consistency of the generated images are crucial, as 3D-inconsistent multi-view images are not a valid descriptor of a scene. However, achieving 3D consistency in 2D diffusion is challenging. Early work~\cite{dreamfusion,hifa,magic3d,i3d23d,in2n} does not alter the diffusion model and relies on consistency derived from NeRF, by directly training NeRF with the inconsistent multi-view images. The NeRF will then converge to an averaged or smoothed version of the scene, according to its model capability, which results in blurred results with few textures and even fails to generate regular patterns like a plaid or checkered pattern. Follow-up work begins to improve the consistency of the diffusion and/or the pipeline. ViCA-NeRF~\cite{vica} achieves consistency by proposing a different pipeline based on key views. ProlificDreamer~\cite{prolificdreamer} makes the diffusion 3D-aware by inputting the camera parameter to the diffusion model and applying per-scene fine-tuning. CSD~\cite{csd}, IVID~\cite{ivid}, and ConsistNet~\cite{consistnet} propose a joint distillation procedure for multiple views, aiming to generate or edit multiple images in one batch consistently, through either attention, depth-based warping, or Kullback–Leibler divergence. 

However, these methods all share two major constraints: (1) the noise used for generation is not controlled, therefore a single view may lead to different and inconsistent generation results with different noises; (2) these methods only study and enforce the consistency between images within a single batch. Nevertheless, the full generation or editing procedure for the scene is across multiple batches, and there might be inconsistencies in different batches. Our \themodel resolves these limitations by proposing novel structured noise and consistency-enforcing training.
\section{\themodel: Methodology}
\label{sec:method}

\begin{figure*}[t!]
\centering
\centerline{\includegraphics[width=1\linewidth]{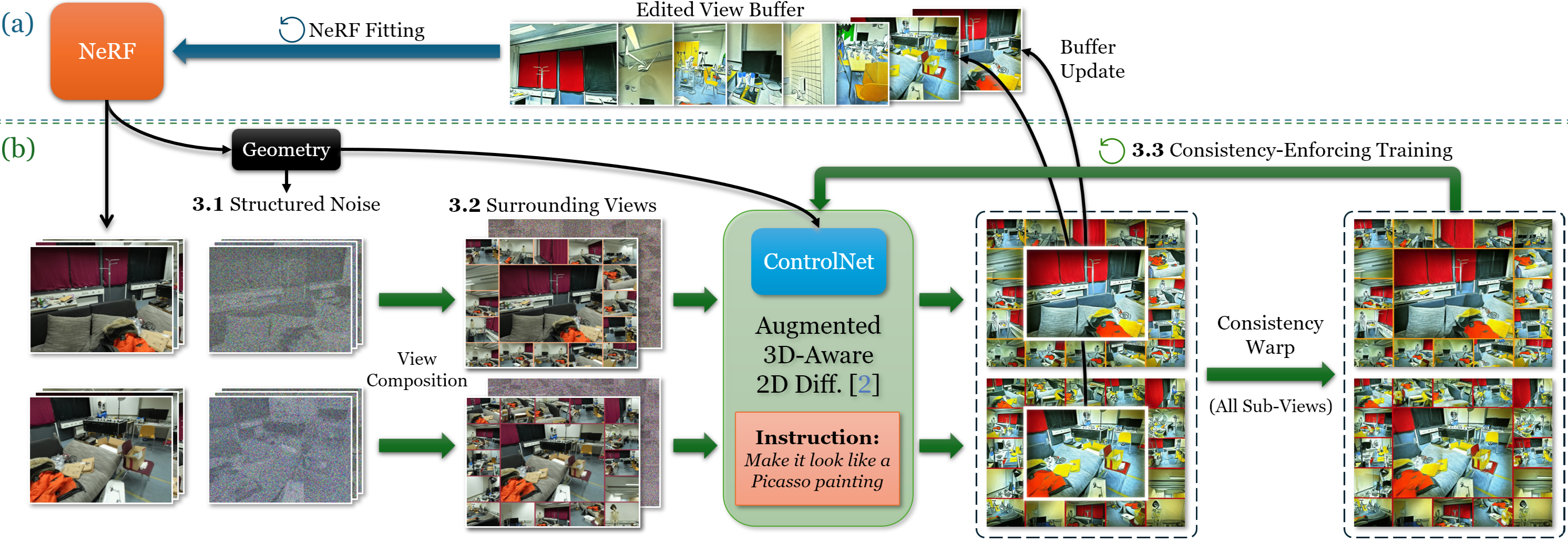}}
\vspace{-3mm}
\caption{\textbf{\themodel framework} is an IN2N-like~\cite{in2n} pipeline containing two major procedures. \textbf{\color{MidnightBlue} (a)} In {\color{MidnightBlue} the NeRF fitting} procedure, we continuously train NeRF with a buffer of edited views. \textbf{\color{ForestGreen} (b)} In {\color{ForestGreen} diffusion generation and training}, we add our 3D-consistent structured noise to rendered multi-view images, and compose surrounding views with them, as input to the augmented 3D-aware 2D diffusion \cite{ip2p}. We then add the edited images to the buffer for {\color{MidnightBlue} (a)}, and apply self-supervised consistency-enforcing training using the consistency-warped images. Note: images of structured noise are only for illustration --  they are actually visually indistinguishable from Gaussian noise images. }
\vspace{-6.5mm}
\label{fig:pipeline}
\end{figure*}

Our \themodel is a novel IN2N-like~\cite{in2n} framework applied upon a diffusion-based 2D image editing model~\cite{ip2p}. As illustrated in Fig.~\ref{fig:pipeline}, our pipeline maintains a buffer of edited views for the NeRF to fit, and uses \cite{ip2p} to generate new edited images for random views according to the instruction, the original appearance, and the current NeRF rendering results. Noticing that the NeRF fitting procedure and diffusion generation procedure are relatively independent, we equivalently execute them in parallel. Within this framework, we propose (1) \emph{structured noise} to enable a 3D-consistent denoising step, starting from 3D-consistent noise at the beginning and ending with 3D-consistent images; (2) \emph{surrounding views} to construct context-rich composed images as input to the 2D diffusion instead of a single view; and (3) a self-supervised \emph{consistency-enforcing training} method via consistent warping in surrounding views, to achieve cross-view and cross-batch consistency.

\subsection{Structured Noise}
\label{sec:method:structnoise}
2D diffusion models generate a new image from a noisy image, which is either pure Gaussian noise or a mixture of noise and the original image. Prior works like DreamFusion~\cite{dreamfusion} and IN2N typically sample different Gaussian noise in each iteration. However, varying noise leads to highly different generation results (as shown in the supplementary material). In other words, previous methods cannot even produce consistent (\ie, identical) images for the same view in different generations, fundamentally limiting their ability to generate consistent results.

This observation motivates us to control and manipulate the noise, by introducing 3D-consistent structured noise. Intuitively, while it is difficult to generate, denoise, or restore 3D-consistent images from inconsistent random noise, the task becomes more manageable when generating consistent images from noise that is itself consistent. 
Therefore, instead of using independently generated noise in each iteration, we generate noise on the surface of the scene \emph{only once} during initialization, and render the noise at each view to obtain the noise used in generating the image for that view. Our strategy aligns with 3D diffusion models like DiffRF~\cite{diffrf}, which directly generate noise in 3D space. The difference lies in the denoising step: while such work directly denoises in 3D, we distill the ``3D denoising process'' from pre-trained 2D diffusion models.

As a latent diffusion model, \cite{ip2p} actually requires noise in latent space, which is $(H/8,W/8,4)$ instead of the image shape $(H,W,3)$. Each element in this noise latent should be independently generated from $N(0,1)$. Constructing such 3D-consistent structured noise remains non-trivial: we need to place noise in 3D, project noise into 2D pixels at multiple scales, and ensure correspondence between different views. Additionally, the distribution of each image's noise should be Gaussian, as noise in an incorrect or dependent distribution may lead to abnormal generation results (as shown in the supplementary material).

To overcome these challenges, we construct a dense point cloud of the scene by unprojecting all the pixels in all the views to points, with the depth predicted by NeRF. For each point $p$, we randomly sample a weighted noise $c(p)=(x,w)$, where $x\sim N(0,1)$ is an independently generated Gaussian noise, and $w\sim U(0,1)$ is its weight. To generate the noise at one view, we identify the sub-point cloud that is front-most in this view, and project it onto the plane. For multiple points projected to the same pixel, we aggregate them by selecting the weighted noise $(x,w)$ with the maximum $w$, and form a noise image $I$ of shape $(H,W)$ consisting of values in $x$. As each $x$ is independently generated and selected (according to $w$), we have $I\sim N(0,1)^{H\times W}$, \ie, making $I$ valid 2D Gaussian noise.

Given that each pixel in the latent space is only roughly related to its corresponding $8\times 8$ region in the image, we can generate noise in the latent space by operating at the downsampled resolution of $(H/8)\times (W/8)$. We thus generate different weighted noise $\{c_i(p)\}$ for each of the four channels of the latent space, and stack the individually rendered noise ${I_i}$ to construct a Gaussian noise image of $(H/8,W/8,4)$, which is then used as the noise by the diffusion model.

The structured noise serves as the foundation for 3D-consistent generation. In Sec.~\ref{sec:method:conft}, we introduce a training method to ensure a consistent denoising procedure from the beginning to the end, so that the denoised images of different views at every denoising step is also 3D consistent.

\subsection{Surrounding Views}
\label{sec:method:surview}
Using the original view as input is a standard practice, when employing 2D diffusion models. This method works well in simple $360^{\circ}$ or forward-facing scenes used by IN2N~\cite{in2n}, as a single view covers most objects in the scene. However, in more complicated scenes like the cluttered rooms in ScanNet++~\cite{scannetpp}, a view may only contain a corner or a bare wall in the room. This hinders the diffusion model to generate plausible results, due to the limited context in a single view.

Intriguingly, our investigation reveals that \cite{ip2p} performs well on composed images, generating an image composed of style-consistent edited sub-images with the same structure (as shown in the supplementary). This observation inspires us to exploit a novel input format for diffusion models -- surrounding views with a composition of one main view and many reference views, so that all views collectively provide contextual information. As illustrated in Fig.~\ref{fig:pipeline}, the key principles in the design of a surrounding view are: (1) the main view that we focus on in this generation should occupy a large proportion; and (2) it should include as many reference views as possible at a reasonable size to provide context. In practice, we construct a surrounding view \wrt a specific main view, by surrounding a large image of this view with $4(k-1)$ small reference images of other views, leaving a margin of arbitrary color. This ensures that the main image is roughly $(k-2)$ times larger than the small images. Here $k$ is a hyperparameter. The reference images are randomly selected from all the views or nearby views from the main view, providing both a global picture of the scene and much overlapped content to benefit training.

We use such surrounding views as input images to \cite{ip2p}, by constructing the surrounding views of the current NeRF's rendering results, structured noise, and original views. Though not directly trained with this image format, \cite{ip2p} still supports generating edited images in the same format, with each image corresponding to the edited result of the image in the same position. The attention modules in its UNet implicitly connect the same regions in different views, enabling the small views to provide extra context to the main view. This results in consistently edited styles among all the sub-images in the surrounding view image.

The surrounding views not only provide a context-rich input format for 2D diffusion models, but also allow it to generate edited results for $(4k\!-\!3)$ views in one batch, benefiting our consistency-enforcing training in Sec.~\ref{sec:method:conft}.

\begin{figure*}[t!]
\centering
\centerline{\includegraphics[width=.87\linewidth]{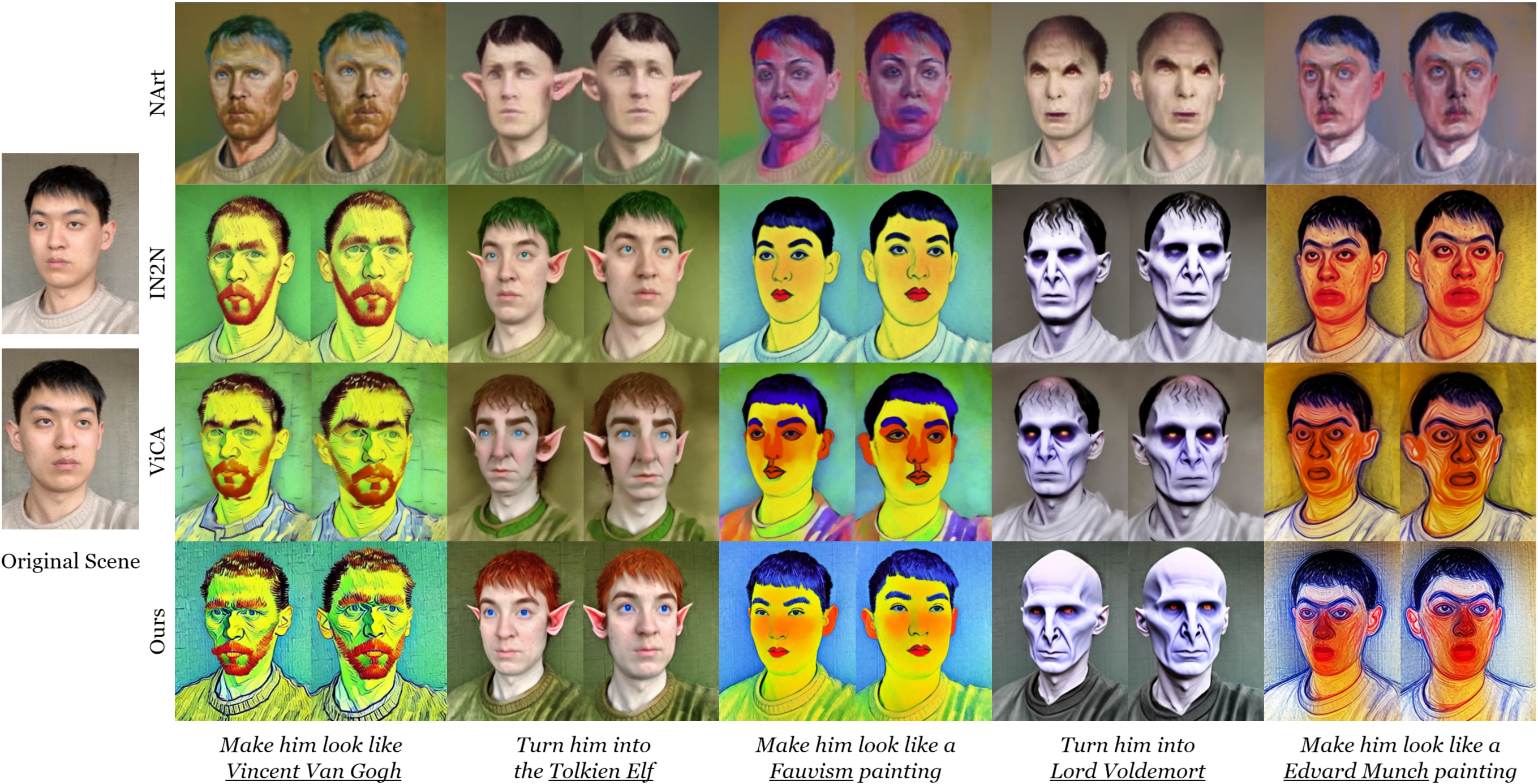}}
\vspace{-3mm}
\caption{\textbf{Comparison in the Fangzhou scene} shows that our \themodel produces significantly sharper editing results with more fine-grained textures and higher consistency with the instruction, \eg, Lord Voldemort with no hair on which all baselines fail. The instructions are the bottom texts, except for NArt \cite{nerfart}, which uses the underlined texts. The images of baselines are taken from their paper.}
\vspace{-6mm}
\label{fig:exp-compare-fangzhou}
\end{figure*}

\subsection{Consistency-Enforcing Training}
\label{sec:method:conft}
We design consistent per-scene training based on structured noise and surrounding views, enforcing 2D diffusion to generate 3D consistent images through a consistent denoising procedure.

\noindent\textbf{Multi-GPU Parallelization Paradigm.} Our pipeline involves training both NeRF and 2D diffusion. Observing that training and inferring a diffusion model is considerably more time-consuming than training the NeRF, while there are very few dependencies between them, we propose a multi-GPU parallelization paradigm. With $(n+1)$ GPUs, we dedicate GPU$0$ to continuously and asynchronously train a NeRF on the buffer of edited images. The remaining $n$ GPUs are utilized to train the diffusion model and generate new edited images added to the buffer for NeRF training. At each diffusion training iteration, we allocate a view to each of the $n$ GPUs and train diffusion on them synchronously. This parallelization eliminates the need to explicitly trade off between NeRF and diffusion training, leading to a \textbf{10$\times$} speed-up in training. With multiple diffusion generations running synchronously, we can also enforce \emph{cross-generation consistency}.

\noindent\textbf{Augmenting 2D Diffusion with 3D-Informing ControlNet.} Intuitively, a 3D-consistent model needs to be 3D-aware; otherwise, it lacks the necessary information and may solely adapt to the input structured noise, potentially leading to overfitting. Therefore, we incorporate an additional ControlNet~\cite{controlnet} adaptor into our 2D diffusion, which injects 3D information as a new condition. The 3D information is obtained by using NeRF to infer the depth and 3D point for each pixel in the view. We then query its feature in a learnable 3D embedding (implemented as a hash table in~\cite{instngp}) to acquire a pixel-wise 3D-aware feature image, which serves as the condition for ControlNet. These components make the augmented diffusion to be aware of, learn from, and generate results based on 3D information. Additionally, we apply LoRA~\cite{lora} to further enhance the capability of diffusion. 

\noindent\textbf{Self-Supervised Consistency Loss.} Lacking ground truth for consistently edited images, we introduce a self-supervised method to enforce 3D consistency. For a set of generated multi-view images, we construct a corresponding reference set of 3D consistent multi-view images to serve as a self-supervision target. Inspired by~\cite{consistnet,ivid}, we employ depth-based warping with NeRF-rendered depth to establish pixel correspondence across views. We design a weighted averaging process to aggregate these pixels to the final image, ensuring multi-view consistency (detail in supplementary).

Specifically, we edit $n$ surrounding views synchronously on $n$ GPUs, with each surrounding view containing $(4k-3)$ views, resulting in a total of $V=(4k-3)n$ views.  For each view $v$, we warp the edited results of the remaining $V-1$ views to it, and compute their weighted average to obtain the reference view $v'$. We then re-aggregate reference views $\{v'\}$ back into surrounding views in the original structure for each GPU. These re-assembled surrounding views are then used as the target images to supervise 2D diffusion.

To guide 2D diffusion in preserving the original style and avoiding smoothing out, we define our \emph{consistency loss} as the sum of the VGG-based perceptual and stylization loss~\cite{vggpercept}, instead of a pixel-wise loss, between diffusion's output and the target image. In addition to this primary loss, we propose several regularization losses to prevent mode collapse and promote 3D awareness (detail in supplementary). With the consistency loss, \themodel effectively enforces not only cross-view consistency among all views in each surrounding view, but also cross-generation or cross-batch consistency for views edited by different GPUs.

\noindent\textbf{Consistent Denoising Procedure.} With our structured noise, the denoising in 2D diffusion initiates with consistent noise. This leads to a further goal to make the {\em entire denoising procedure} 3D consistent and thus end with consistent images. We achieve this by enforcing all the views in the intermediate denoising images to be also 3D consistent at each denoising step. Therefore, unlike the conventional diffusion training with single-step denoising, our training involves a full multi-step denoising procedure with passing through gradients. As it is impossible to fit the entire computational graph into the GPU memory, we use checkpointing~\cite{checkpointing1,checkpointing2} to trade space with time. Doing so enables constructing the reference set of images with warping for each intermediate denoising step, which is then used to supervise the intermediate denoising image. This provides more direct signals of 3D consistency in the training of diffusion, facilitating the generation of 3D consistent results.

\noindent\textbf{Shape Editing.} Some instructions, \eg, \textit{Make him smile}, change the shape or geometry of the scene during editing, while our structured noise and consistency-enforcing training rely on the geometry. To be compatible with shape editing, we design a \emph{coarse-to-fine} strategy: we first edit the scene using \themodel with only the surrounding view and disabling the other two components, \ie, using image-independent noise and the original implementation of \cite{ip2p}. This allows the scene to converge to a coarse edited shape according to the instruction. We then activate structured noise and consistency-enforcing training to refine the editing. We periodically adjust the structured noise with changes in geometry, while preserving the noise values. With this strategy, \themodel also achieves high-fidelity shape editing. 

\section{Experiments}
\label{sec:expr}

\paragraph{Editing Tasks.} In our setting, each editing task is a pair of $(\texttt{scene},\texttt{instruction})$, indicating which instruction-guided editing operation should be applied on which scene. The output of the task is another scene, being the edited scene under the instruction. The scenes we use for evaluation contain two parts:
(1) {\em IN2N.} Scenes used by IN2N~\cite{nerf}, including scenes of human faces or bodies, outdoor scenes, and statues; and (2) {\em SN++.} Scenes in ScanNet++~\cite{scannetpp}, which are complicated indoor scenes with free-formed structures and camera trajectories. We also use two types of editing instructions: (1) {\em style transfer} which transfers the style of the scene into the described style, and (2) {\em object-specific editing} which edits a specific object of the scene. We use these tasks to compare our approach with baselines, and conduct ablation study on representative tasks.

\paragraph{NeRF Backbone and Diffusion Model.} 
For a fair comparison 
with previous works~\cite{in2n,vica}, we use the Nerfacto model in NeRFStudio~\cite{nerfstudio} as our NeRF backbone, and the pre-trained diffusion model \cite{ip2p} from Hugging Face as our initial checkpoint. The NeRF representation for the scene is trained with NeRFStudio in advance, and then used in our pipeline.

\paragraph{\themodel Variants.} We investigate the following variants for our ablation study (where $-$SN, $-$SV, and $-$T denote removing structured noise, surrounding views, and consistency-enforcing training, respectively): (1) Full \themodel. (2) No structured noise ($-$SN): use independently generated noise for each view instead of structured noise, but still use surrounding views and perform consistency-enforcing training. (3) No training ($-$T): use surrounding views and structured noise, but do not augment and train \cite{ip2p} and keep using the original checkpoint. (4) Only surrounding views ($-$SN $-$T): only use surrounding views, and do not use structured noise or train \cite{ip2p}. (5) ``IN2N'' ($-$SN $-$SV $-$T): ours with all the proposed components removed, which can be regarded as an alternative version of IN2N. Note that consistency-enforcing training requires surrounding views to produce sufficient edited views in one generation; we cannot remove surrounding views but still apply consistency-enforcing training on \cite{ip2p}.

\paragraph{Baselines.} We mainly compare our method with two baselines: Instruct-NeRF2NeRF (IN2N)~\cite{in2n} and ViCA-NeRF (ViCA)~\cite{vica}, as they are most closely related to our task. We also compare with NeRF-Art (NArt)~\cite{nerfart} as an early work. Other methods, however, lack publicly available or working code and/or only use a few scenes supported by NerfStudio. Therefore, we could only compare with CSD~\cite{csd}, DreamEditor~\cite{dreameditor}, GE \cite{gausseditor}, EN2N \cite{en2n}, and PDS \cite{pds} under a few tasks in supplementary, and are unable to compare with Edit-DiffNeRF~\cite{editdiffnerf} and Instruct 3D-to-3D~\cite{i3d23d}. Note that \themodel solves instruction-guided scene editing instead of scene generation, so we do not compare with models for the generation task~\cite{dreamfusion,consistnet,ivid,hifa}.

\begin{table}[t!]
\centering
\centerline{\includegraphics[width=1\linewidth]{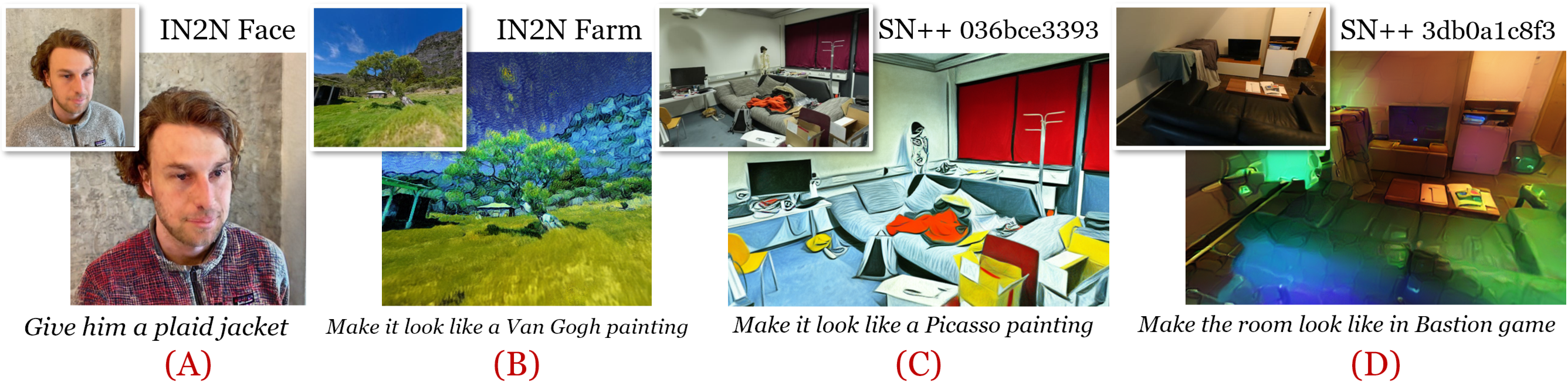}}
\scalebox{0.68}{\begin{tabular}{ll|cccc}
 \hline\hline
 Variant & Components & A & B & C & D \\
 \hline 
 Full & All                     &$\textbf{27.4}^{\pm0.4}$ & $\textbf{159.8}^{\pm2.1}$ & $\textbf{62.7}^{\pm1.9}$ & $\textbf{132.8}^{\pm1.2}$ \\
 \hline 
 No Str. Noise & $-$SN          &$35.0^{\pm0.5}$ &$234.8^{\pm2.4}$ &$83.1^{\pm2.2}$ &$217.8^{\pm2.0}$ \\
 No Training & $-$T             &$34.6^{\pm1.1}$ &$221.8^{\pm2.0}$ &$80.4^{\pm2.0}$ &$201.8^{\pm1.9}$ \\
 Only Sur. Views & $-$SN $-$T   &$34.0^{\pm0.3}$ &$262.0^{\pm2.7}$ &$83.9^{\pm2.2}$ &$214.4^{\pm2.0}$ \\
 \hline 
 ``IN2N'' & $-$SN $-$SV $-$T    &$91.0^{\pm1.2}$ &$255.8^{\pm2.0}$ &$90.9^{\pm1.5}$ &$222.4^{\pm2.0}$ \\
 \hline\hline
\end{tabular}}
\vspace{-3mm}
\caption{\textbf{Ablation study} on the distillation fidelity score ($\downarrow$) \emph{quantitatively} validates the effectiveness and complementarity of each of our components. Our full \themodel significantly outperforms all variants across various scenes and types of instructions.}
\vspace{-6.5mm}
\label{tab:exp-ablation}
\end{table}

\paragraph{Evaluation Metrics.} Observing that our \themodel generates significantly sharper editing results, consistent with previous work~\cite{in2n,vica}, we compare \themodel with baselines mainly through qualitative evaluation. For the ablation study, the appearance of the scenes edited by our different variants may be visually similar and unable to be fairly compared using qualitative results. Therefore, we propose \emph{distillation fidelity score} (DFS) to evaluate how faithful the editing is distilled and applied on NeRF compared with the diffusion's output \cite{ip2p}, rooted in the basic setting that we distill from \cite{ip2p} to edit 3D scenes. In this situation, our editing ability is bounded by \cite{ip2p}'s. Consistent with the training objective of DreamFusion~\cite{dreamfusion}, we aim to minimize the distance between two distributions: the distribution of a rendered image at a random view from the edited NeRF, and the distribution of the diffusion editing result of an image at a random view in the original scene.
Following this, we define the fidelity metric as the Fréchet inception distance (FID)~\cite{fid,fidpytorch} between two sets -- the set of images rendered by the edited NeRF at all training views, and the set of edited images generated by the original \cite{ip2p} for all training views, corresponding to these two distributions. A lower FID means a higher fidelity that the editing is applied to the scene.

\begin{figure*}[t!]
\centering

\scalebox{1}{
\noindent 
\begin{minipage}[t]{0.78432\textwidth}
    \vspace{0pt}
    \centerline{\includegraphics[width=1\linewidth]{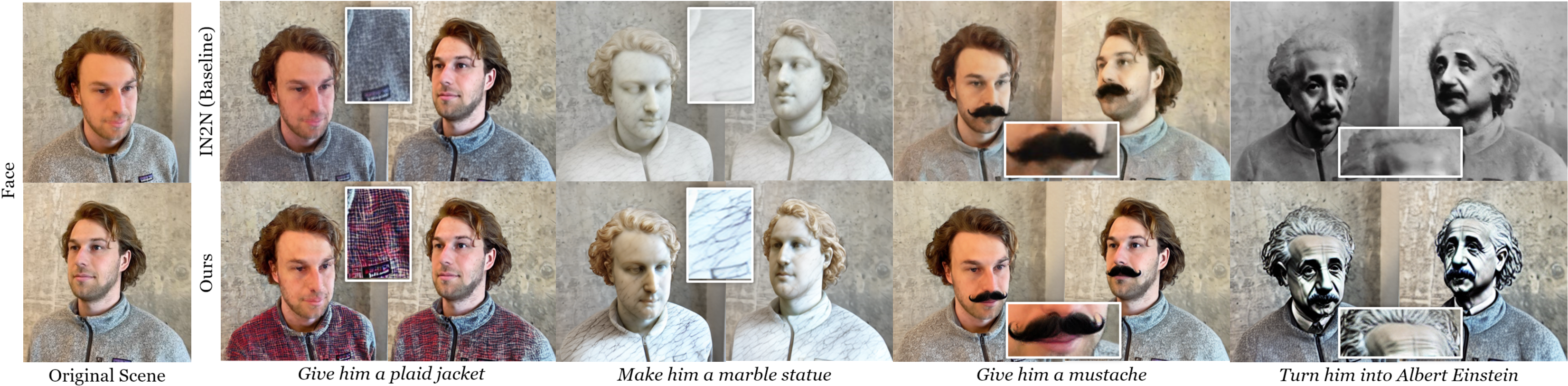}}
    \centerline{\includegraphics[width=1\linewidth]{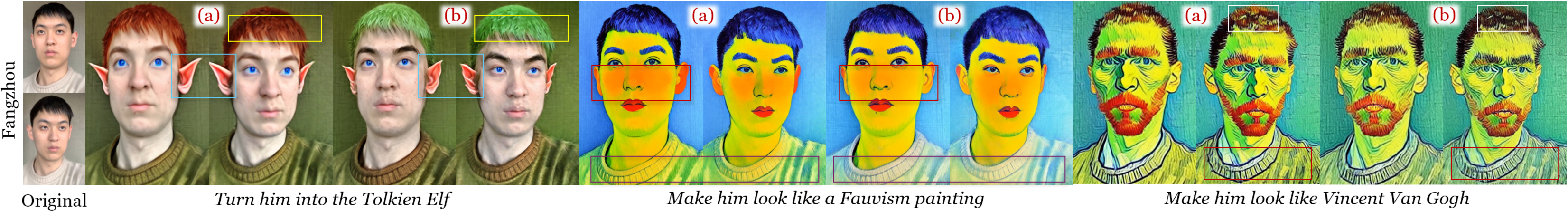}}

    \centerline{\includegraphics[width=1\linewidth]{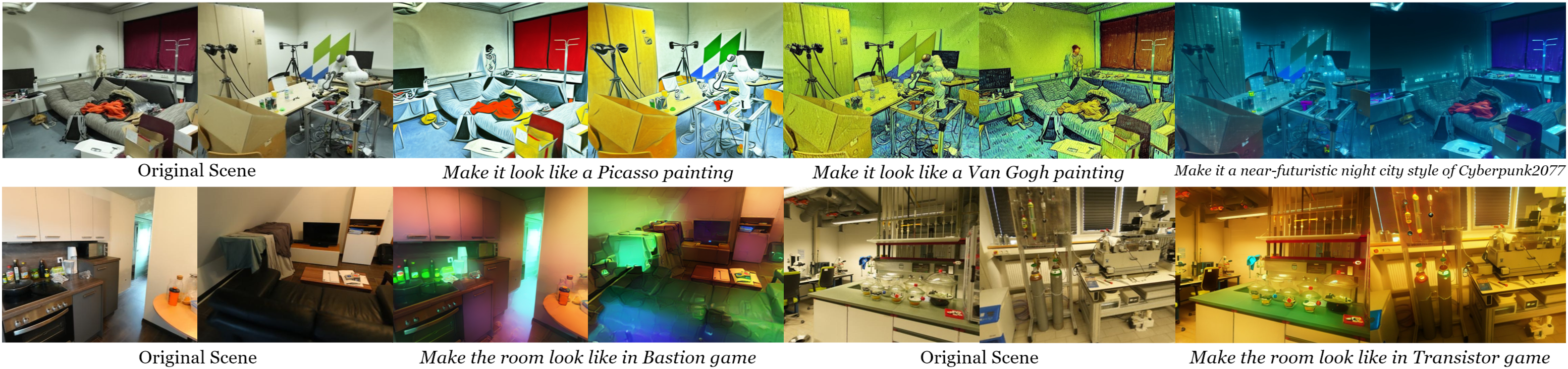}}
\end{minipage}\begin{minipage}[t]{0.21568\textwidth}
    \vspace{0pt}
    \centerline{\includegraphics[width=1\linewidth]{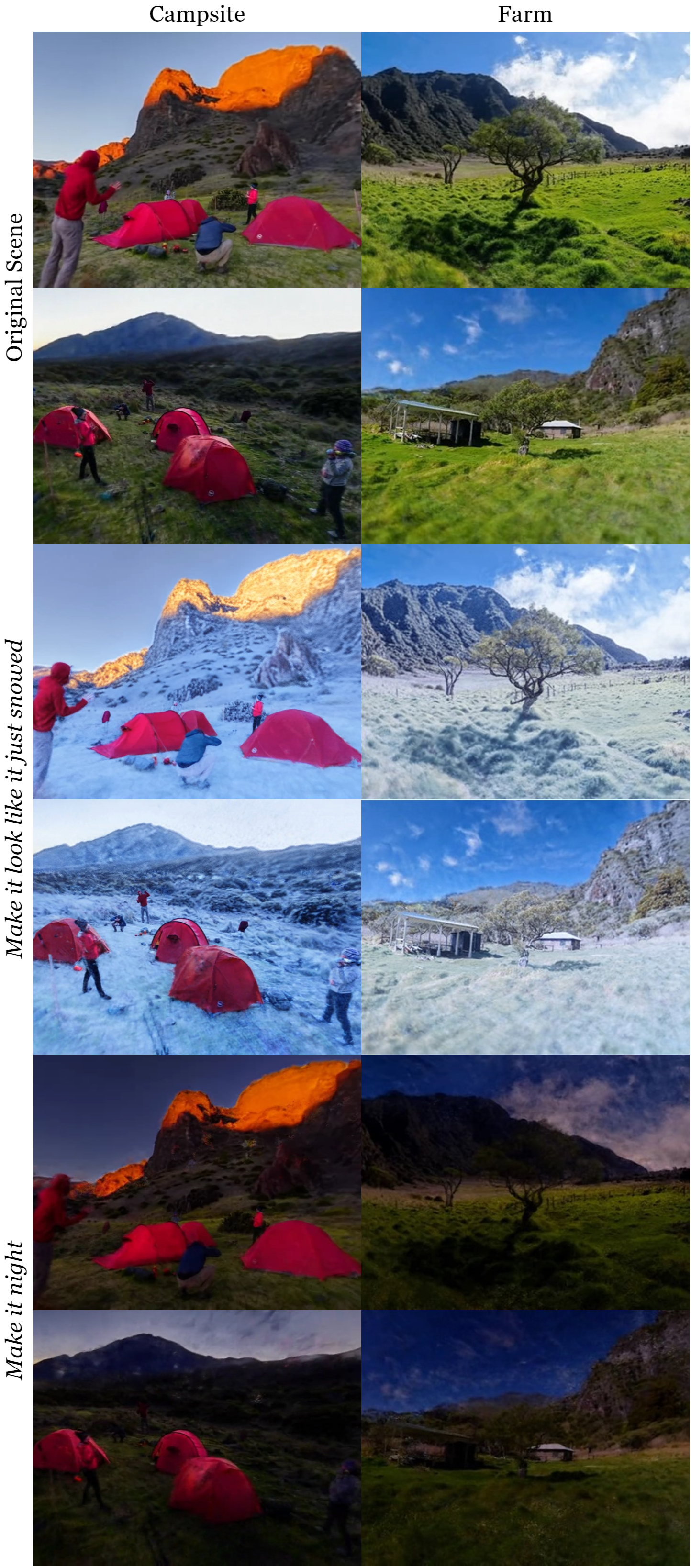}}
\end{minipage}
}

\vspace{-3mm}
\caption{\themodel \emph{consistently} generates high-quality and high-fidelity editing results, featuring detailed, fine-grained textures across various scenes and instructions. Notably, \themodel also maintains the high diversity from \cite{ip2p}, as exemplified by the highly diversified results \textcolor[RGB]{192,0,0}{(a)(b)}. \textbf{Additional results and comparisons are provided in the supplementary and on our project page}.}
\vspace{-6.5mm}
\label{fig:exp-gallery}
\end{figure*}

\paragraph{Qualitative Results.} The qualitative comparison in the Fangzhou scene from the IN2N dataset is shown in Fig.~\ref{fig:exp-compare-fangzhou}. Distilling from the same diffusion model \cite{ip2p}, IN2N~\cite{in2n}, ViCA~\cite{vica}, and our \themodel produce results in a similar style. As especially shown in the ``Vincent Van Gogh'' and ``Edvard Munch'' editing, our \themodel generates results containing fine-grained representative textures of Van Gogh and Munch, while the baseline results are blurred and only contain simple or coarse textures. This validates that with our proposed components, \themodel is able to generate consistent images from \cite{ip2p} with detailed textures, and does not rely on consistency derived from NeRF, which, unfortunately, smooths out the results. Notably, in the ``Lord Voldemort'' case, our \themodel is the \emph{only one} that successfully edits the image to resemble the well-known, distinctive appearance of Lord Voldemort, featuring no hair and a peculiar nose. Among all the editing tasks, our \themodel consistently produces editing results with the most detailed ears and hair/head, and does not contain unnatural color blocks.

Additional qualitative results are shown in Fig.~\ref{fig:exp-gallery}, and more results and the comparison with baselines on these tasks are provided in the supplementary and on our project page. Overall, our \themodel generates sharp, bright editing results in all tasks across various scenes, including human, indoor, and outdoor scenes. 
(1) In the Face scene, our \themodel successfully applies the plaid (checkered) jacket editing, a common failure case in most previous methods, including IN2N. Also, our \themodel is able to assign fine-grained marble texture in the marble statue editing, a clear mustache in Mustache editing, and clear wrinkle and hair in Einstein editing, while IN2N produces blurred and over-smooth results with poor details. Notably, our \themodel minimizes the side effects of the editing, while IN2N unexpectedly and significantly changes the skin color in the Mustache editing and the wall color in the Einstein editing.
(2) The Tolkien Elf and Fauvism editing tasks in the Fangzhou scene show that our \themodel could preserve most diversity from the original \cite{ip2p}, due to the use of structured noise sampled for the whole editing. With the structured noise, we can focus on the consistency of generation for the given noise, without suffering from averaging results generated from different noises, which may lose diversity by converging to an average style for all noises. 
(3) Our \themodel works well in outdoor scenes, as all the details on the floor, mountain, plants, and camps are preserved in the edited results. 
(4) In complicated indoor scenes from the ScanNet++ dataset, our \themodel generates editing results that are easy to recognize as the given style, with fine-grained textures (Van Gogh), regular patterns (Picasso), or special lighting conditions (Bastion and Transistor). 
All these results validate that our \themodel generates high-quality editing results.

\paragraph{Ablation Study.} As shown in Table~\ref{tab:exp-ablation}, we conduct the ablation study on four representative tasks \textcolor[RGB]{192,0,0}{(A)}-\textcolor[RGB]{192,0,0}{(D)}, covering instructions of object-specific editing, artistic style transfer, and other style transfer, and scenes of human, indoor, and outdoor scenes.  The results show that our full \themodel outperforms all the variants with significant gains in all tasks under DFS, which mainly comes from our consistent denoising procedure in Sec. \ref{sec:method:conft} that requires all three major components to achieve. Training towards a consistent denoising procedure produces considerable extra supervision signals to the augmented \cite{ip2p}, making it converge better towards consistent generation results. We can also observe that the consistency-enforcing training and the use of surrounding views improve the fidelity in most of the tasks, especially in the complicated large-scale indoor scenes \textcolor[RGB]{192,0,0}{(C)(D)}, showing that these components indeed improve the consistency in generation.

\section{Conclusion}
\label{sec:conclusion}
This paper proposes \themodel, an instruction-guided scene editing framework that generates 3D consistently edited images from 2D diffusion models. Empirical evaluation shows that \themodel produces editing results of significantly higher quality, exhibiting sharper, brighter appearance with fine-grained textures, across various scenes including forward-facing human scenes, outdoor scenes, and even large-scale indoor scenes in ScanNet++, where it succeeds in common failure cases of previous methods. We hope that our work can serve as a source of inspiration for distillation-based 3D/4D editing and generation tasks.

\noindent{\footnotesize\textbf{Acknowledgement.} Jun-Kun and Yu-Xiong were supported in part by NSF Grant 2106825 and NIFA Award 2020-67021-32799, using NVIDIA GPUs at NCSA Delta through allocations CIS220014 and CIS230012 from the ACCESS program.}

{
    \small
    \bibliographystyle{ieeenat_fullname}
    \bibliography{main}

\begin{thebibliography}{49}
\providecommand{\natexlab}[1]{#1}
\providecommand{\url}[1]{\texttt{#1}}
\expandafter\ifx\csname urlstyle\endcsname\relax
  \providecommand{\doi}[1]{doi: #1}\else
  \providecommand{\doi}{doi: \begingroup \urlstyle{rm}\Url}\fi

\bibitem[Barron et~al.(2021)Barron, Mildenhall, Tancik, Hedman, Martin-Brualla, and Srinivasan]{mipnerf}
Jonathan~T. Barron, Ben Mildenhall, Matthew Tancik, Peter Hedman, Ricardo Martin-Brualla, and Pratul~P. Srinivasan.
\newblock {Mip-{NeRF}}: A multiscale representation for anti-aliasing neural radiance fields.
\newblock In \emph{ICCV}, 2021.

\bibitem[Brooks et~al.(2023)Brooks, Holynski, and Efros]{ip2p}
Tim Brooks, Aleksander Holynski, and Alexei~A. Efros.
\newblock Learning to follow image editing instructions.
\newblock In \emph{CVPR}, 2023.

\bibitem[Chen et~al.(2021)Chen, Xu, Zhao, Zhang, Xiang, Yu, and Su]{mvsnerf}
Anpei Chen, Zexiang Xu, Fuqiang Zhao, Xiaoshuai Zhang, Fanbo Xiang, Jingyi Yu, and Hao Su.
\newblock {MVS{NeRF}}: Fast generalizable radiance field reconstruction from multi-view stereo.
\newblock In \emph{ICCV}, 2021.

\bibitem[Chen et~al.(2023{\natexlab{a}})Chen, Lyu, and Wang]{neuraleditor}
Jun-Kun Chen, Jipeng Lyu, and Yu-Xiong Wang.
\newblock {NeuralEditor}: Editing neural radiance fields via manipulating point clouds.
\newblock In \emph{CVPR}, 2023{\natexlab{a}}.

\bibitem[Chen et~al.(2023{\natexlab{b}})Chen, Chen, Zhang, Wang, Yang, Wang, Cai, Yang, Liu, and Lin]{gausseditor}
Yiwen Chen, Zilong Chen, Chi Zhang, Feng Wang, Xiaofeng Yang, Yikai Wang, Zhongang Cai, Lei Yang, Huaping Liu, and Guosheng Lin.
\newblock {GaussianEditor}: Swift and controllable 3d editing with gaussian splatting, 2023{\natexlab{b}}.

\bibitem[Dong and Wang(2023)]{vica}
Jiahua Dong and Yu-Xiong Wang.
\newblock {ViCA-{NeRF}}: View-consistency-aware {3D} editing of neural radiance fields.
\newblock In \emph{NeurIPS}, 2023.

\bibitem[Guo et~al.(2022)Guo, Kang, Bao, He, and Zhang]{nerfren}
Yuan-Chen Guo, Di Kang, Linchao Bao, Yu He, and Song-Hai Zhang.
\newblock {{NeRF}ReN}: Neural radiance fields with reflections.
\newblock In \emph{CVPR}, 2022.

\bibitem[Haque et~al.(2023)Haque, Tancik, Efros, Holynski, and Kanazawa]{in2n}
Ayaan Haque, Matthew Tancik, Alexei Efros, Aleksander Holynski, and Angjoo Kanazawa.
\newblock {Instruct-{NeRF}2{NeRF}}: Editing {3D} scenes with instructions.
\newblock In \emph{ICCV}, 2023.

\bibitem[Heusel et~al.(2017)Heusel, Ramsauer, Unterthiner, Nessler, and Hochreiter]{fid}
Martin Heusel, Hubert Ramsauer, Thomas Unterthiner, Bernhard Nessler, and Sepp Hochreiter.
\newblock {GAN}s trained by a two time-scale update rule converge to a local nash equilibrium.
\newblock In \emph{NeurIPS}, 2017.

\bibitem[Ho et~al.(2020)Ho, Jain, and Abbeel]{diffusion}
Jonathan Ho, Ajay Jain, and Pieter Abbeel.
\newblock Denoising diffusion probabilistic models.
\newblock In \emph{NeurIPS}, 2020.

\bibitem[Hu et~al.(2022)Hu, Shen, Wallis, Allen-Zhu, Li, Wang, Wang, and Chen]{lora}
Edward~J Hu, Yelong Shen, Phillip Wallis, Zeyuan Allen-Zhu, Yuanzhi Li, Shean Wang, Lu Wang, and Weizhu Chen.
\newblock Lo{RA}: Low-rank adaptation of large language models.
\newblock In \emph{ICLR}, 2022.

\bibitem[Johnson et~al.(2016)Johnson, Alahi, and Fei-Fei]{vggpercept}
Justin Johnson, Alexandre Alahi, and Li Fei-Fei.
\newblock Perceptual losses for real-time style transfer and super-resolution.
\newblock In \emph{ECCV}, 2016.

\bibitem[Kamata et~al.(2023)Kamata, Sakuma, Hayakawa, Ishii, and Narihira]{i3d23d}
Hiromichi Kamata, Yuiko Sakuma, Akio Hayakawa, Masato Ishii, and Takuya Narihira.
\newblock {Instruct {3D}-to-{3D}}: Text instruction guided {{3D}-to-{3D}} conversion.
\newblock \emph{arXiv preprint arXiv:2303.15780}, 2023.

\bibitem[Kerbl et~al.(2023)Kerbl, Kopanas, Leimk{\"u}hler, and Drettakis]{3dgs}
Bernhard Kerbl, Georgios Kopanas, Thomas Leimk{\"u}hler, and George Drettakis.
\newblock 3d gaussian splatting for real-time radiance field rendering.
\newblock \emph{TOG}, 2023.

\bibitem[Kim et~al.(2023)Kim, Lee, Choi, Jeong, Sohn2, and Shin1]{csd}
Subin Kim, Kyungmin Lee, June~Suk Choi, Jongheon Jeong, Kihyuk Sohn2, and Jinwoo Shin1.
\newblock Collaborative score distillation for consistent visual editing.
\newblock In \emph{NeurIPS}, 2023.

\bibitem[Kobayashi et~al.(2022)Kobayashi, Matsumoto, and Sitzmann]{distillnerf}
Sosuke Kobayashi, Eiichi Matsumoto, and Vincent Sitzmann.
\newblock Decomposing {{NeRF}} for editing via feature field distillation.
\newblock In \emph{NeurIPS}, 2022.

\bibitem[Koo et~al.(2023)Koo, Park, and Sung]{pds}
Juil Koo, Chanho Park, and Minhyuk Sung.
\newblock Posterior distillation sampling, 2023.

\bibitem[Lin et~al.(2023)Lin, Gao, Tang, Takikawa, Zeng, Huang, Kreis, Fidler, Liu, and Lin]{magic3d}
Chen-Hsuan Lin, Jun Gao, Luming Tang, Towaki Takikawa, Xiaohui Zeng, Xun Huang, Karsten Kreis, Sanja Fidler, Ming-Yu Liu, and Tsung-Yi Lin.
\newblock Magic{3D}: High-resolution text-to-{3D} content creation.
\newblock In \emph{CVPR}, 2023.

\bibitem[Liu et~al.(2020)Liu, Gu, Lin, Chua, and Theobalt]{neuralsparse}
Lingjie Liu, Jiatao Gu, Kyaw~Zaw Lin, Tat-Seng Chua, and Christian Theobalt.
\newblock Neural sparse voxel fields.
\newblock In \emph{NeurIPS}, 2020.

\bibitem[Liu et~al.(2021)Liu, Zhang, Zhang, Zhang, Zhu, and Russell]{editnerf}
Steven Liu, Xiuming Zhang, Zhoutong Zhang, Richard Zhang, Junyan Zhu, and Bryan~C. Russell.
\newblock Editing conditional radiance fields.
\newblock In \emph{ICCV}, 2021.

\bibitem[Mildenhall et~al.(2020)Mildenhall, Srinivasan, Tancik, Barron, Ramamoorthi, and Ng]{nerf}
Ben Mildenhall, Pratul~P. Srinivasan, Matthew Tancik, Jonathan~T. Barron, Ravi Ramamoorthi, and Ren Ng.
\newblock {{NeRF}}: Representing scenes as neural radiance fields for view synthesis.
\newblock In \emph{ECCV}, 2020.

\bibitem[M{\"u}ller et~al.(2023)M{\"u}ller, Siddiqui, Porzi, Bulo, Kontschieder, and Nie{\ss}ner]{diffrf}
Norman M{\"u}ller, Yawar Siddiqui, Lorenzo Porzi, Samuel~Rota Bulo, Peter Kontschieder, and Matthias Nie{\ss}ner.
\newblock {DiffRF}: Rendering-guided {3D} radiance field diffusion.
\newblock In \emph{CVPR}, 2023.

\bibitem[M\"uller et~al.(2022)M\"uller, Evans, Schied, and Keller]{instngp}
Thomas M\"uller, Alex Evans, Christoph Schied, and Alexander Keller.
\newblock Instant neural graphics primitives with a multiresolution hash encoding.
\newblock \emph{ACM Trans. Graph.}, 41\penalty0 (4):\penalty0 102:1--102:15, 2022.

\bibitem[Peng et~al.(2022)Peng, Yan, Liu, Cheng, Guan, Pan, Zhai, and Yang]{cagenerf}
Yicong Peng, Yichao Yan, Shengqi Liu, Yuhao Cheng, Shanyan Guan, Bowen Pan, Guangtao Zhai, and Xiaokang Yang.
\newblock Cage{{NeRF}}: Cage-based neural radiance field for generalized {{3D}} deformation and animation.
\newblock In \emph{NeurIPS}, 2022.

\bibitem[Poole et~al.(2023)Poole, Jain, Barron, and Mildenhall]{dreamfusion}
Ben Poole, Ajay Jain, Jonathan~T. Barron, and Ben Mildenhall.
\newblock {DreamFusion}: Text-to-{3D} using {2D} diffusion.
\newblock In \emph{ICLR}, 2023.

\bibitem[Radford et~al.(2021)Radford, Kim, Hallacy, Ramesh, Goh, Agarwal, Sastry, Askell, Mishkin, Clark, Krueger, and Sutskever]{clip}
Alec Radford, Jong~Wook Kim, Chris Hallacy, Aditya Ramesh, Gabriel Goh, Sandhini Agarwal, Girish Sastry, Amanda Askell, Pamela Mishkin, Jack Clark, Gretchen Krueger, and Ilya Sutskever.
\newblock Learning transferable visual models from natural language supervision.
\newblock In \emph{ICML}, 2021.

\bibitem[Rojas et~al.(2020)Rojas, Kahira, Meneses, Bautista{-}Gomez, and Badia]{checkpointing1}
Elvis Rojas, Albert~Njoroge Kahira, Esteban Meneses, Leonardo Bautista{-}Gomez, and Rosa~M. Badia.
\newblock A study of checkpointing in large scale training of deep neural networks.
\newblock \emph{arXiv preprint arXiv:2012.00825}, 2020.

\bibitem[Rombach et~al.(2022)Rombach, Blattmann, Lorenz, Esser, and Ommer]{sd}
Robin Rombach, Andreas Blattmann, Dominik Lorenz, Patrick Esser, and Björn Ommer.
\newblock High-resolution image synthesis with latent diffusion models.
\newblock In \emph{CVPR}, 2022.

\bibitem[Ruiz et~al.(2023)Ruiz, Li, Jampani, Pritch, Rubinstein, and Aberman]{dreambooth}
Nataniel Ruiz, Yuanzhen Li, Varun Jampani, Yael Pritch, Michael Rubinstein, and Kfir Aberman.
\newblock Dreambooth: Fine tuning text-to-image diffusion models for subject-driven generation.
\newblock In \emph{CVPR}, 2023.

\bibitem[Seitzer(2020)]{fidpytorch}
Maximilian Seitzer.
\newblock {pytorch-fid: {FID} Score for PyTorch}.
\newblock \url{https://github.com/mseitzer/pytorch-fid}, 2020.
\newblock Version 0.3.0.

\bibitem[Song et~al.(2023)Song, Cao, Gu, Jiang, Yuan, and Tang]{en2n}
Liangchen Song, Liangliang Cao, Jiatao Gu, Yifan Jiang, Junsong Yuan, and Hao Tang.
\newblock {Efficient-NeRF2NeRF}: Streamlining text-driven 3d editing with multiview correspondence-enhanced diffusion models, 2023.

\bibitem[Tancik et~al.(2023)Tancik, Weber, Ng, Li, Yi, Kerr, Wang, Kristoffersen, Austin, Salahi, Ahuja, McAllister, and Kanazawa]{nerfstudio}
Matthew Tancik, Ethan Weber, Evonne Ng, Ruilong Li, Brent Yi, Justin Kerr, Terrance Wang, Alexander Kristoffersen, Jake Austin, Kamyar Salahi, Abhik Ahuja, David McAllister, and Angjoo Kanazawa.
\newblock {Nerfstudio}: A modular framework for neural radiance field development.
\newblock In \emph{SIGGRAPH}, 2023.

\bibitem[Verbin et~al.(2022)Verbin, Hedman, Mildenhall, Zickler, Barron, and Srinivasan]{refnerf}
Dor Verbin, Peter Hedman, Ben Mildenhall, Todd Zickler, Jonathan~T. Barron, and Pratul~P. Srinivasan.
\newblock {Ref-{NeRF}}: Structured view-dependent appearance for neural radiance fields.
\newblock In \emph{CVPR}, 2022.

\bibitem[Wang et~al.(2023{\natexlab{a}})Wang, Jiang, Chai, He, Chen, and Liao]{nerfart}
Can Wang, Ruixiang Jiang, Menglei Chai, Mingming He, Dongdong Chen, and Jing Liao.
\newblock {{NeRF}-Art}: Text-driven neural radiance fields stylization.
\newblock \emph{IEEE Transactions on Visualization and Computer Graphics}, pages 1--15, 2023{\natexlab{a}}.

\bibitem[Wang et~al.(2021)Wang, Wang, Genova, Srinivasan, Zhou, Barron, Martin-Brualla, Snavely, and Funkhouser]{ibrnet}
Qianqian Wang, Zhicheng Wang, Kyle Genova, Pratul Srinivasan, Howard Zhou, Jonathan~T. Barron, Ricardo Martin-Brualla, Noah Snavely, and Thomas Funkhouser.
\newblock {IBRNet}: Learning multi-view image-based rendering.
\newblock In \emph{CVPR}, 2021.

\bibitem[Wang et~al.(2023{\natexlab{b}})Wang, Lu, Wang, Bao, Li, Su, and Zhu]{prolificdreamer}
Zhengyi Wang, Cheng Lu, Yikai Wang, Fan Bao, Chongxuan Li, Hang Su, and Jun Zhu.
\newblock {ProlificDreamer}: High-fidelity and diverse text-to-{3D} generation with variational score distillation.
\newblock In \emph{NeurIPS}, 2023{\natexlab{b}}.

\bibitem[Xiang et~al.(2023)Xiang, Yang, Huang, and Tong]{ivid}
Jianfeng Xiang, Jiaolong Yang, Binbin Huang, and Xin Tong.
\newblock {{3D}}-aware image generation using {{2D}} diffusion models.
\newblock In \emph{ICCV}, 2023.

\bibitem[Xu et~al.(2021)Xu, Xu, Philip, Bi, Shu, Sunkavalli, and Neumann]{pointnerf}
Qiangeng Xu, Zexiang Xu, Julien Philip, Sai Bi, Zhixin Shu, Kalyan Sunkavalli, and Ulrich Neumann.
\newblock {Point-{NeRF}}: Point-based neural radiance fields.
\newblock In \emph{CVPR}, 2021.

\bibitem[Xu and Harada(2022)]{deformingnerf}
Tianhan Xu and Tatsuya Harada.
\newblock Deforming radiance fields with cages.
\newblock In \emph{ECCV}, 2022.

\bibitem[Xu et~al.(2022)Xu, Liu, Tao, Xuan, and Zhang]{checkpointing2}
Xiangzhe Xu, Hongyu Liu, Guanhong Tao, Zhou Xuan, and Xiangyu Zhang.
\newblock Checkpointing and deterministic training for deep learning.
\newblock In \emph{2022 IEEE/ACM 1st International Conference on AI Engineering – Software Engineering for AI (CAIN)}, 2022.

\bibitem[Yang et~al.(2021)Yang, Zhang, Xu, Li, Zhou, Bao, Zhang, and Cui]{objectnerf}
Bangbang Yang, Yinda Zhang, Yinghao Xu, Yijin Li, Han Zhou, Hujun Bao, Guofeng Zhang, and Zhaopeng Cui.
\newblock Learning object-compositional neural radiance field for editable scene rendering.
\newblock In \emph{ICCV}, 2021.

\bibitem[Yang et~al.(2023)Yang, Cheng, Duan, Ji, and Li]{consistnet}
Jiayu Yang, Ziang Cheng, Yunfei Duan, Pan Ji, and Hongdong Li.
\newblock {ConsistNet}: Enforcing {3D} consistency for multi-view images diffusion.
\newblock \emph{arXiv preprint arXiv:2310.10343}, 2023.

\bibitem[Yeshwanth et~al.(2023)Yeshwanth, Liu, Nie{\ss}ner, and Dai]{scannetpp}
Chandan Yeshwanth, Yueh-Cheng Liu, Matthias Nie{\ss}ner, and Angela Dai.
\newblock {ScanNet++}: A high-fidelity dataset of {3D} indoor scenes.
\newblock In \emph{ICCV}, 2023.

\bibitem[Yu et~al.(2021)Yu, Ye, Tancik, and Kanazawa]{pixelnerf}
Alex Yu, Vickie Ye, Matthew Tancik, and Angjoo Kanazawa.
\newblock {pixel{NeRF}}: Neural radiance fields from one or few images.
\newblock In \emph{CVPR}, 2021.

\bibitem[Yu et~al.(2023)Yu, Xiang, and Han]{editdiffnerf}
Lu Yu, Wei Xiang, and Kang Han.
\newblock {Edit-DiffNeRF}: Editing {3D} neural radiance fields using {2D} diffusion model.
\newblock \emph{arXiv preprint arXiv:2306.09551}, 2023.

\bibitem[Yuan et~al.(2022)Yuan, tian Sun, Lai, Ma, Jia, and Gao]{nerfediting}
Yu-Jie Yuan, Yang tian Sun, Yu-Kun Lai, Yuewen Ma, Rongfei Jia, and Lin Gao.
\newblock {{NeRF}-Editing}: Geometry editing of neural radiance fields.
\newblock In \emph{CVPR}, 2022.

\bibitem[Zhang et~al.(2023)Zhang, Rao, and Agrawala]{controlnet}
Lvmin Zhang, Anyi Rao, and Maneesh Agrawala.
\newblock Adding conditional control to text-to-image diffusion models.
\newblock In \emph{ICCV}, 2023.

\bibitem[Zhu and Zhuang(2023)]{hifa}
Joseph Zhu and Peiye Zhuang.
\newblock {HiFA}: High-fidelity text-to-{3D} with advanced diffusion guidance.
\newblock \emph{arXiv preprint arXiv:2305.18766}, 2023.

\bibitem[Zhuang et~al.(2023)Zhuang, Wang, Liu, Lin, and Li]{dreameditor}
Jingyu Zhuang, Chen Wang, Lingjie Liu, Liang Lin, and Guanbin Li.
\newblock {DreamEditor}: Text-driven {{3D}} scene editing with neural fields.
\newblock In \emph{SIGGRAPH Asia}, 2023.

\end{thebibliography}
}

\twocolumn[{%
\maketitle
\renewcommand\twocolumn[1][]{#1}%
    \centering
    \Large
    \vspace{-1em}\textbf{Supplementary Material} \\
    \vspace{1.0em}
}]

\appendix

This document contains additional analysis and extra experiments. The content of this document is summarized as below:

\renewcommand{\thetable}{\Alph{section}.\arabic{subsection}}
\renewcommand{\thefigure}{\Alph{section}.\arabic{subsection}}

\setcounter{tocdepth}{2}
\etocsetlevel{section}{1}
\def\authcount{}
\etocsettocstyle{\color{white}}{}
\renewcommand{\contentsname}{}
\localtableofcontents

\counterwithin{figure}{section}
\counterwithin{table}{section}

\section{\color{DarkOrange} Supplementary Video (\sv)}
\label{sec:supplvideo}
To better visualize our results and compare with baselines beyond static 2D images, we provide a \textcolor{DarkOrange}{supplementary video (\sv)} on our project page at \href{https://immortalco.github.io/ConsistDreamer/}{\textit{immortalco.github.io/ConsistDreamer}}. 
We also include a short demo in this video, to enhance the understanding of 3D-consistent structured noise. The original size of the video is around 1.25GB, therefore we have to compress it to fit it in the upload size limitation of 200MB on OpenReview.

In the following sections, we use \sv to refer to this supplementary video. 

\section{Comparisons with Additional Baselines}

\begin{figure}[t]
    \centering
    %
    \centerline{\includegraphics[width=1\linewidth]{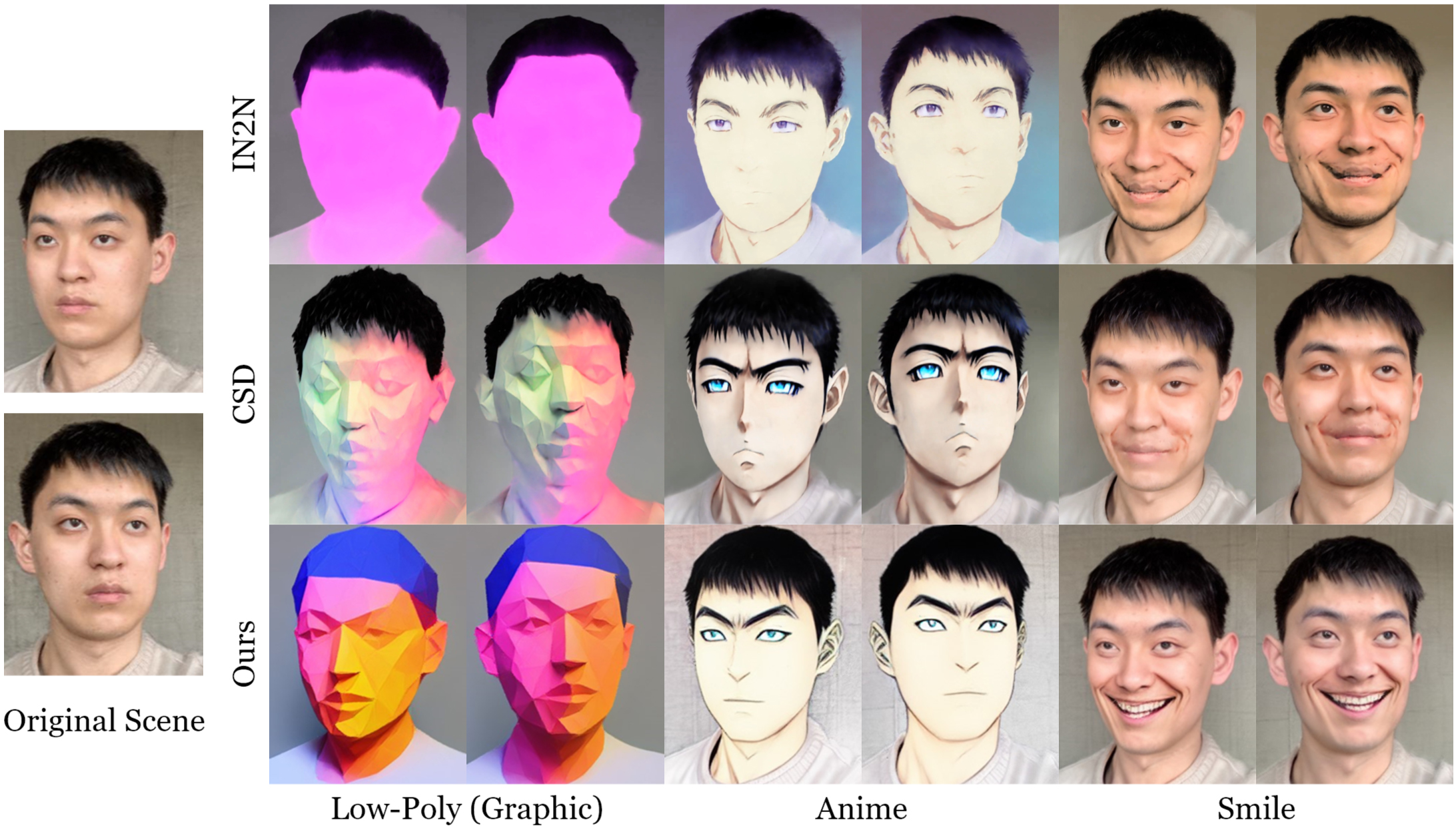}}
    
    \caption{Qualitative comparisons with baseline CSD on three tasks show that our \themodel achieves high-quality editing, outperforming both IN2N and CSD with more successful editing.   }
    
    \label{fig:exp-suppl-csd}
\end{figure}

\begin{figure}[t]
    \centering
    \centerline{\includegraphics[width=1\linewidth]{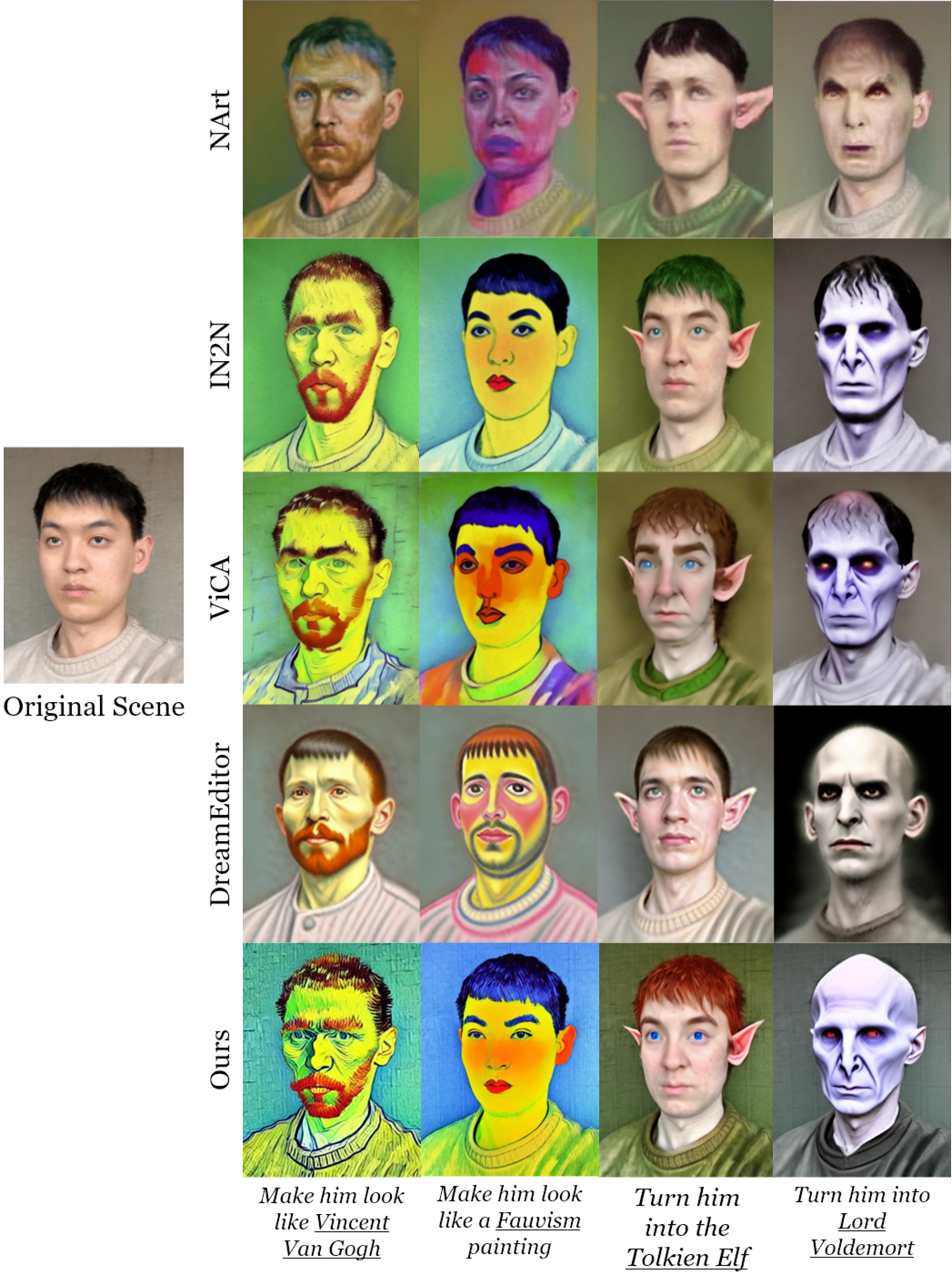}}
    
    \caption{Compared with DreamEditor, our \themodel achieves better editing, which not only follows and satisfies the given instructions, but also preserves as much content of the original scene as possible. On the contrary, DreamEditor completely edits the original person to another in all the tasks. }
    
    \label{fig:exp-suppl-dreameditor}
\end{figure}

\begin{figure}[t]
    \centering
    \centerline{\includegraphics[width=1\linewidth]{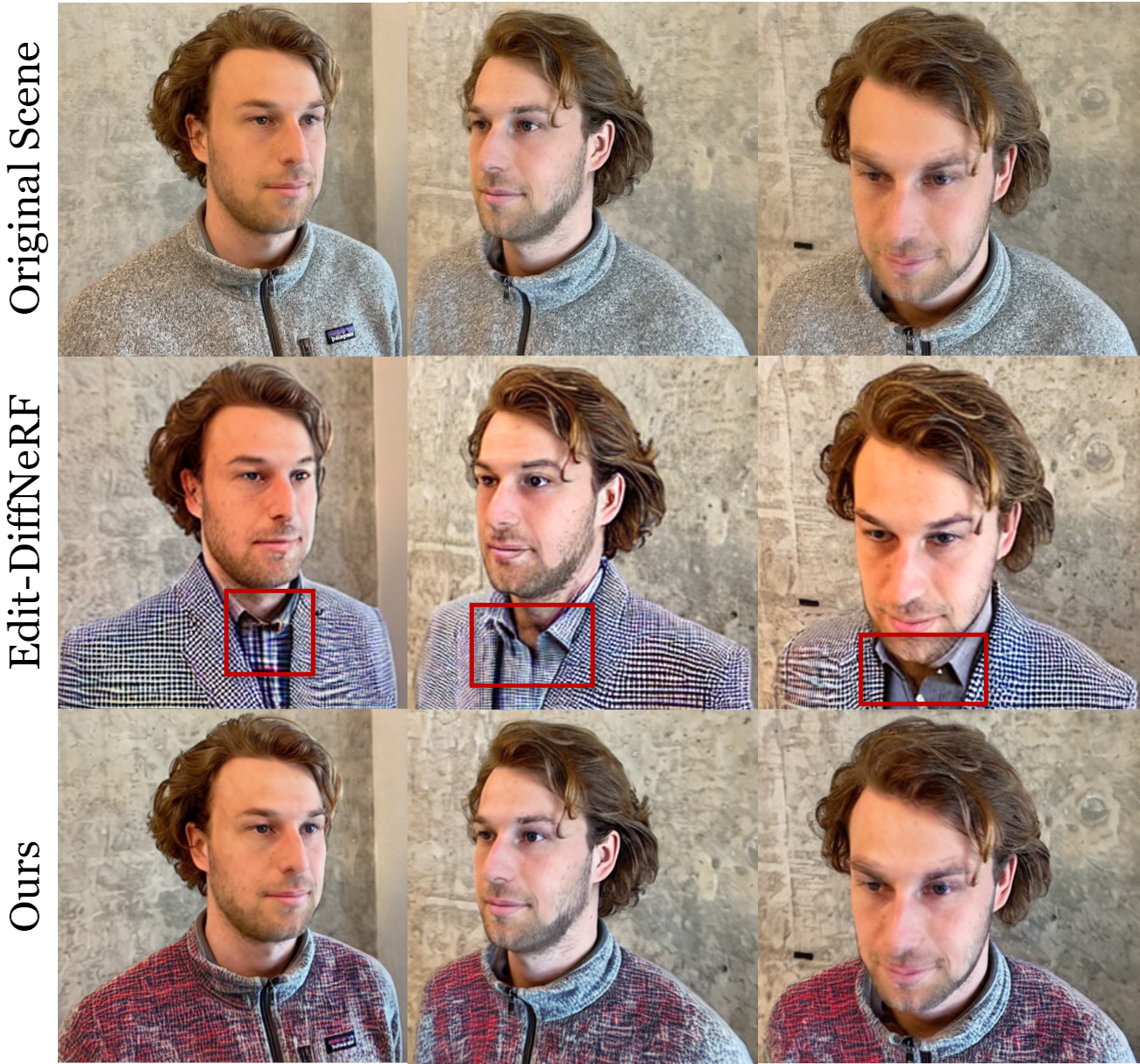}}
    
    \caption{Our \themodel achieves consistent editing in the checkered/plaid pattern (also visualized as smooth video in \svat{3:01-3:16}), while Edit-DiffNeRF has obvious inconsistency in the shape and texture of the collar.}
    
    \label{fig:exp-suppl-editdiffnerf}
\end{figure}
In the main paper, we compare our \themodel with IN2N~\cite{in2n} and ViCA~\cite{vica}. In this section, we compare our \themodel with other baselines and provide some analysis. These methods either do not have publicly available code, or evaluate on the scenes which are not supported by NeRFStudio. Therefore, we could only compare our \themodel under the tasks used by them, with the provided visualizations from their papers or websites.

We also provide some comparisons in the video format of the baselines in \sv.
\subsection{CSD~\cite{csd}}
CSD is a method focusing on general consistent generation, including large image editing, scene editing, and scene generation. We compare our \themodel with CSD under three tasks shown on \href{https://subin-kim-cv.github.io/CSD/}{the website of CSD}\footnote{https://subin-kim-cv.github.io/CSD/}: Low-Poly (Graphic), Anime, and Smile. 

As shown in Fig. \ref{fig:exp-suppl-csd} and \svat{03:30-03:38}, our \themodel significantly outperforms IN2N, which fails in the Low-Poly and Anime tasks, and has the side effects of adding beards in the Smile task. Compared with CSD, our editing in the Low-Poly task is more noticeable, with a successfully edited hair part. Our edited scene in the Smile task is the only one among all three to successfully show the teeth when smiling, while CSD's result contains strange muscles as if the person is keeping a straight face. In conclusion, our \themodel achieves more successful editing than CSD.

\subsection{DreamEditor~\cite{dreameditor}}
DreamEditor is another method focusing on scene editing, but with another diffusion model \cite{dreambooth} instead of \cite{ip2p}. As NeRFStudio does not support the other scenes, we compare our \themodel with DreamEditor by comparing Fig. 3 in our main paper with Fig. 8 in \cite{dreameditor}.

Fig. \ref{fig:exp-suppl-dreameditor} presents the results in these tasks, along with other baselines in Fig. 3 in our main paper. It shows that our \themodel preserves most of the contents in the original scene while editing, \eg, the shape of the head and face, and the shape and type of the clothes, minimizing the side effects of editing. DreamEditor, however, completely edits the person to another person, even in the Fauvism task, which is supposed to be only style transfer. This demonstrates that our \themodel achieves more reasonable editing than DreamEditor.

\subsection{Edit-DiffNeRF~\cite{editdiffnerf}}
Edit-DiffNeRF is another paper that also claims to successfully complete the checkered/plaid pattern. As they did not provide any code, we compare our \themodel with the images provided in their paper. As shown in Fig. \ref{fig:exp-suppl-editdiffnerf}, our \themodel achieves consistent editing among all three views, while Edit-DiffNeRF's results are multi-view inconsistent, obviously shown in the collar part. The smooth video of our rendering result in \svat{3:01-3:16} also shows the consistency of our \themodel. These results validate that our \themodel archives significantly better consistency in checkered/plaid patterns, while Edit-DiffNeRF fails to achieve such consistency.

\subsection{Instruct 3D-to-3D~\cite{i3d23d}}
Instruct 3D-to-3D is a method focusing on style transfer of scenes. It uses LLFF and NeRF Synthetic (NS) scenes as editing tasks instead of the widely-used IN2N dataset. 
In contrast, we focus on editing more challenging and realistic scenes.
In addition, as NeRFStudio and NeRFacto do not support LLFF and NS datasets well (more specifically, NeRFStudio does not support the LLFF dataset, and NeRFacto works well in real scenes but not in synthetic scenes like NS), we cannot compare with Instruct 3D-to-3D on these two datasets. Moreover, the code of Instruct 3D-to-3D is not publicly available. Therefore, we are unable to compare with Instruct 3D-to-3D.

\subsection{Concurrent Works: GE \cite{gausseditor}, EN2N \cite{en2n}, And PDS \cite{pds}}

\begin{figure*}[t]
    \centering
    %
    \centerline{\includegraphics[width=1\linewidth]{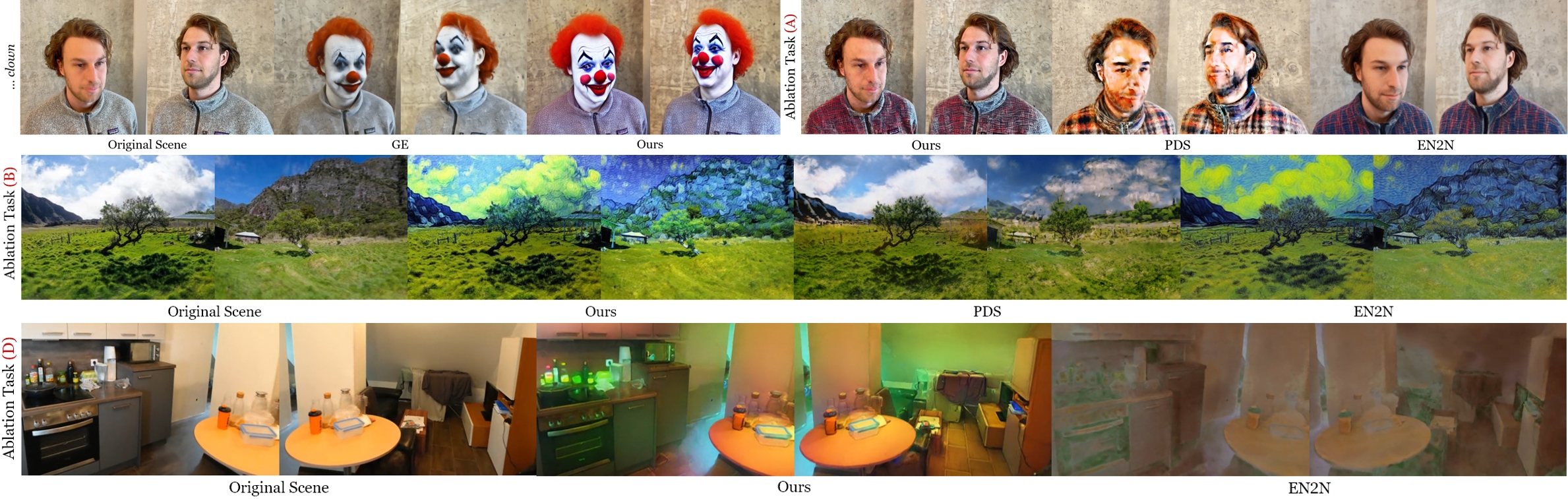}}
    
    \caption{Our \themodel outperforms all three concurrent works with brighter color, clearer textures, and better editing results matched with the editing instruction. }
    
    \label{fig:exp-concur}
\end{figure*}

\cite{gausseditor,pds,en2n} are three concurrent works. GE \cite{gausseditor} and EN2N \cite{en2n} achieve 3D editing through the same 2D diffusion model \cite{ip2p} and have some modifications in the pipeline or scene representation, while PDS \cite{pds} proposes another distillation formula and uses DreamBooth \cite{dreambooth} for editing.

The comparisons against them are in Fig. \ref{fig:exp-concur}. Our \themodel generates high-quality editing results with brighter color and clearer textures, while all these concurrent works generate blurred textures, gloomy colors, and/or unsuccessful or unreasonable editing.

\section{CLIP \cite{clip} Metrics In IN2N \cite{in2n}}

\begin{table*}[t!]
\centering
\scalebox{1}{\begin{tabular}{l|cccc|cccc}
 \hline\hline
 
 \multirow{2}{*}{Method} &
 \multicolumn{4}{c|}{\small Text-Image Direction Similarity (CTIDS) $\uparrow$} &
 \multicolumn{4}{c|}{\small Direction Consistency (CDC) $\uparrow$}
 \\ 
 \cline{2-9}
 & A & B & C & D & A & B & C & D\\
 \hline 
Ours & \textbf{0.0259} & \textbf{0.1679} & \textbf{0.1204} & \textbf{0.1268} & \textbf{0.5785} & \textbf{0.1735} & \textbf{0.3077} & \textbf{0.2878}\\
IN2N \cite{in2n} & 0.0099 & 0.1252 & 0.1163 & 0.1055 & 0.5106 & 0.1634 & 0.2900 & 0.1772\\
 \hline\hline
\end{tabular}}
\caption{Our \themodel significantly and consistently outperforms baseline IN2N in CLIP \cite{clip} metrics over all four ablation scenes.}

\label{tab:suppl-clip}
\end{table*}

We provide the quantitative comparison with CLIP \cite{clip} metrics introduced in IN2N \cite{in2n} in Tab. \ref{tab:suppl-clip}. In all four ablation scenes, ours significantly and consistently outperforms IN2N in both metrics. 

\section{Implementation Details}
\begin{figure*}[t!]
\centering
\centerline{\includegraphics[width=1\linewidth]{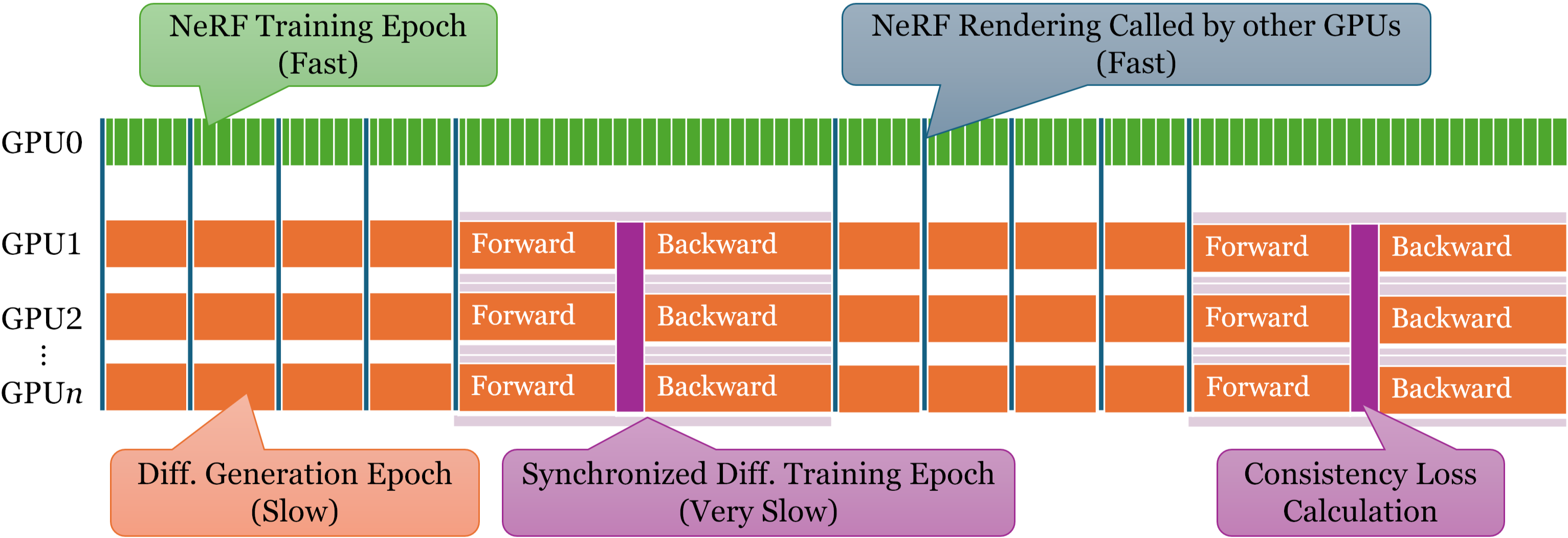}}

\caption{Our multi-GPU training pipeline uses $n+1$ GPUs, one dedicated for NeRF fitting, and the others for diffusion training and generation. }

\label{fig:multigpu}
\end{figure*}

\subsection{Hyperparameters and Settings}
In our experiments, we use the multi-GPU pipeline with $n=3$ (4 GPUs in total), and surrounding views of $k=5$ (1 main view and 12 reference views). 

The learning rate of each component is shown below:
\begin{itemize}
    \item NeRFacto \cite{nerfstudio}: $5\times 10^{-3}$ for field part, and $10^{-2}$ for proposal network part.
    \item LoRA-augmented diffusion model \cite{ip2p}: consistent with their original implementation ($10^{-4}$).
    \item Learnable 3D positional embedding: $2\times 10^{-3}$.
\end{itemize}

All the views are resized to $3:4$ or $4:3$ according to their orientations. For landscape images (portrait images use the same setting with a flipped height and width), the diffusion model takes a surrounding view image input at $1152\times 864$ (also $4:3$), with horizontal splitters at heights $6$pix, and vertical splitters at widths $8$pix. The sizes of the main view and reference views are $688\times 516$ and $224\times 168$, respectively. This setting is consistent with the original usage of diffusion models \cite{ip2p,sd} trained at $512\times 512$, as our main view has a height close to it.

Consistent with IN2N \cite{in2n}, both MSE and LPIPS losses are used to train NeRF. 

\subsection{Viewpoints And Camera Trajectory}

During the distillation process, we directly use the viewpoints provided in the original scene dataset, which is sufficient to cover the whole scene. In visualization, we use the provided camera trajectory for IN2N \cite{in2n} dataset, and manually construct another camera trajectory for a smooth visualization for ScanNet++ \cite{scannetpp} dataset.

\subsection{Training Schedule}
\label{sec:supp-impl-train}
One standard full training contains $1,600$ epochs across multiple sub-stages. All these stages are explained below:
\begin{itemize}
\item Initialization Stage (Epoch $1\sim200$): Train diffusion \cite{ip2p} before NeRF fitting. We perform one diffusion training step in one epoch.
    \begin{itemize}
        \item Early Bootstrap (Epoch $1\sim50$): Train the LoRA-augmented diffusion model to mimic the behavior of the original model with the augmented input of 3D positional embedding. The weight regularization loss of maintaining original behavior (detailed in \ref{sec:supp-impl-reg}) is significantly higher. NeRF training has not started.
        \item Bootstrap (Epoch $51\sim 150$): Train the consistency-awareness of the LoRA-augmented diffusion model while keeping original behavior, at a similar importance with balanced weights. 
        \item Warming Up (Epoch $151\sim 200$): Use the standard weights to balance the consistency loss and regularization, focusing more on consistency. This epoch generates sufficient images for the edited view buffer for NeRF fitting.
    \end{itemize}
\item Distillation Stage (Epoch $201\sim 1600$): Train diffusion while fitting NeRF. In each of 4 epochs, we do 3 diffusion generation steps without training (to fill the edited view buffer), and only one diffusion training step. Here, the ``noise level'' means the mixture rate of the current NeRF (being edited) rendered image and the noise as the diffusion's input: full noise level means using only noise for generation (standard generation), while a $30\%$ noise level means the input image is the mixture of $30\%$ noise and $70\%$ rendered image.
    \begin{itemize}
    \item Full Noise Generation (Epoch $201\sim500$): The diffusion model is trained and used for generation at a full noise level to edit the views sufficiently regardless of NeRF.
    \item Pre-Annealing (Epoch $501\sim600$): The diffusion model is trained and used for generation with a noise level sampled from $[70\%,100\%]$. It edits the views with a few references to the current NeRF, starting to refine the current NeRF.
    \item Annealing (Epoch $601\sim 1500$): Following the idea of HiFA \cite{hifa}, the range of the noise level linearly anneals from $[70\%,100\%]$ to $[10\%,40\%]$. The NeRF will gradually converge to a fine-grained edited version. 
    \item Ending (Epoch $1501\sim 1600$): The diffusion model is trained and used for generation with a noise level sampled from the annealed range $[10\%,40\%]$, to further refine the edited NeRF.
    \end{itemize}
\end{itemize}

If the editing task requires editing the geometry or shape of the scene (``shape editing''), the depth-based warping using the depth of the original scene will be inaccurate. Therefore, in the Initialization Stage, we put the original diffusion model's output to the edited view buffer for NeRF fitting, equivalently using IN2N in this stage. In the distillation stage, the shape of the NeRF will be adjusted to the edited shape in a short time, and then we will start to use the trained diffusion model's output for NeRF fitting.

In our experiments, most IN2N scenes converge to a fine-grained edited scene at $600\sim 700$ epochs, while ScanNet++ \cite{scannetpp} scenes take around $1000$ epochs. 

\subsection{Structured Noise Implementation}
In the main paper, the structured noise is implemented by constructing ``a dense point cloud of the scene by unprojecting all the pixels in all the views'', and rendering/projecting such a point cloud at a view to generate the structured noise. Directly implementing this literal description is complicated and inefficient. 

Therefore, we use an equivalent implementation. 
\begin{itemize}
    \item Instead of explicitly generating this dense point cloud, we just put the weighted noises on each pixel of all views.
    \item For the view we query for structured noise, we warp the noise from all other views to it. This is equivalent to projecting the sub-point cloud generated by each view to the querying view; therefore, it is equivalent to the original design.
\end{itemize}

With this implementation, explicitly generating, maintaining, and projecting a point cloud with billions of points (number of views $\times$ height $\times$ width) is unnecessary, and a query can be completed in less than one second.

\subsection{Surrounding Views - Reference View Selection}
We construct the surrounding view with one large main view and several small reference views. The purpose of the reference views is two-folded: (1) to provide enough context about the whole scene, and (2) to have enough overlapped parts of the main view to facilitate consistency-enforcing training. Therefore, we select $40\%$ of the views to be a random view of the scene, and the rest $60\%$ of the views to be a view with at least $20\%$ overlap of the main view (quantified by the area of matched pixels through warping). The order of the views is randomly shuffled. We observed that none of these randomnesses highly alter the editing result -- after consistency-enforcing training, any choice of reference views and their order will lead to a consistent edited result of the main view.
\subsection{Training - Pixel Weights}
In consistency-enforcing training, we apply warping and weighted averages to compute the training reference views $\{v'\}$, so that all the views in $\{v'\}$ are 3D consistent. Using identical weights for all pixels will result in blurred images: In a scene of a person, one view only contains their face, and another view contains the whole body. Warping the latter to the former indicates an upsampling of the face part, which will be blurred. Merging the blurred, warped view with the former view at the same weight results in blurred overall results.

We propose a better pixel-weighting strategy based on a further analysis of this situation. If we warp pixel $a$ to pixel $b$, where $a$ has a larger ``scope'' and contains more scene objects, then we need to upsample the $b$ part of the view from $a$, resulting in blurry. Therefore, the weight should be related to the scope of the pixel. Following this, we define the pixel area to quantify this scope. For a pixel $p$ in a view from camera position $o$, the four vertices of the pixel grids correspond to the rays $\{o+td_i\}_{i=1}^4$. We use NeRF to predict each of their depth $\{t_i\}_{i=1}^4$, and calculate their corresponding points $P_i = o + t_i d_i$. The pixel area $S(p)$ of this pixel is defined as the area of a square with vertices $P_1,P_2,P_3,P_4$ in the 3D space, which can be regarded as an approximation of the surface area the pixel represents. As we need a lower weight for a pixel with a larger scope, \textit{i.e.,} larger $S(p)$, we define the weight as $1/S(p)$, which satisfies all our needs.

\subsection{Training - Multi-GPU Pipeline}

An illustration of our multi-GPU training pipeline is in Fig. \ref{fig:multigpu}. By implementing such a parallelization pipeline by ourselves, we decouple NeRF training with diffusion generation and training in the most asynchronized way, waiving the necessity of trade-offs between NeRF training and diffusion, achieving considerable speed up.

\subsection{Training - Regularizations}
\label{sec:supp-impl-reg}
We use the consistency loss as the main loss in the consistency-enforcing training. However, this loss only enforces several equalities (required by consistency), leading to trivial results of a pure-color image without regularization losses -- this is also reasonable as all pure-color images of the same color are perfectly consistent. Also, there is no encouragement or enforcement to use the 3D information in the 3D positional embedding. To avoid these, we propose several regularization losses, as shown below:
\begin{itemize}
    \item Maintain Original Behavior. We expect that the trained diffusion model will generate images that are very similar to the original model when all the inputs (image, noises, and 3D positional embeddings) are identical. Therefore, we use MSE and VGG perceptual and stylization losses, to regularize both the generated images and the constructed referenced images (with gradient, generated by warping and averaging) of the trained diffusion, with the original model's output. We further expect that the UNet in the trained diffusion model predicts similar noises at each denoising step as the original UNet, so we also use this to regularize during each denoising step.
    \item Encourage 3D Information Utilization. The original \cite{ip2p} takes the original image of the scene as another part of input, using it as a condition to generate the edited image. To encourage 3D information utilization, we design a regularization loss, to enforce the diffusion model \emph{without} the original image input to generate very similar results to the one \emph{with} the original image input (both with 3D positional embedding input). With the lack of the original image, the only way for the diffusion model to perceive the original view is the 3D positional embedding. Therefore, the diffusion is trained to use the 3D positional embedding at least for novel view synthesis to recover the original image, encouraging the utilization of 3D information. This regularization loss is also applied on the UNet in each denoising step.
    \item Encourage Consistent Editing Style. The diffusion model has some diversity in editing. However, we need to converge to one specific style in one editing procedure, otherwise, the NeRF may use view-dependency to overfit different styles at different views. Therefore, in the Pre-Annealing step (Sec. \ref{sec:supp-impl-train}), we use the NeRF's rendering result to supervise the diffusion model, to make it converge to the style NeRF converges to.
\end{itemize}

\subsection{Variant ``IN2N'' And IN2N \cite{in2n}}
In our ablation study in the main paper, we have a variant ``IN2N'' being our full \themodel with all three major components removed. In this section, we discuss how it is equivalent to an implementation of IN2N, and the major differences between them.

IN2N is a method that (1) gradually generates newly edited images with a noise level (detailed in Sec. \ref{sec:supp-impl-train}) sampled from $[70\%,98\%]$, and (2) uses the newly generated images to fit the NeRF, while the fitting NeRF's rendering results can affect the following editing (through the input of diffusion model as a mixture with noise). This matches our pre-annealing sub-stage. Therefore, ``IN2N'' includes vanilla IN2N as a sub-procedure. Additionally, ``IN2N'' has the following improvements beyond IN2N:
\begin{itemize}
    \item IN2N only samples noise levels from $[70\%,98\%]$. This makes IN2N (1) sometimes unable to sufficiently edit the scene due to the absence of 100\% noise level editing (\eg, unable to achieve a Lord Voldemort editing with no hair in Fig. \ref{fig:exp-suppl-dreameditor}), and (2) cannot refine the editing results based on a converged style, and sometimes even deviates from a converged style to another, as the noise level is always as high as $70\%$. The variant ``IN2N'' starts at a full noise before the pre-annealing sub-stage, guaranteeing sufficient editing. After the pre-annealing sub-stage, ``IN2N'' anneals the noise level range to refine the results, leading to a more fine-grained editing.
    \item IN2N adds the newly edited image to the dataset by replacing a subset of pixels, which may negatively affect the LPIPS/perceptual loss. ``IN2N'' uses an edited view buffer to fit NeRF containing only full, edited views, on which the perceptual loss can perform well.
\end{itemize}
In conclusion, our variant, ``IN2N,'' is an equivalent and improved implementation of IN2N. As shown in \svat{4:52-5:22}, ``IN2N'' generates noticeably better results than IN2N.

\section{Supporting Evidence for Claims}
\begin{figure*}[t!]
\centering
\centerline{\includegraphics[width=1\linewidth]{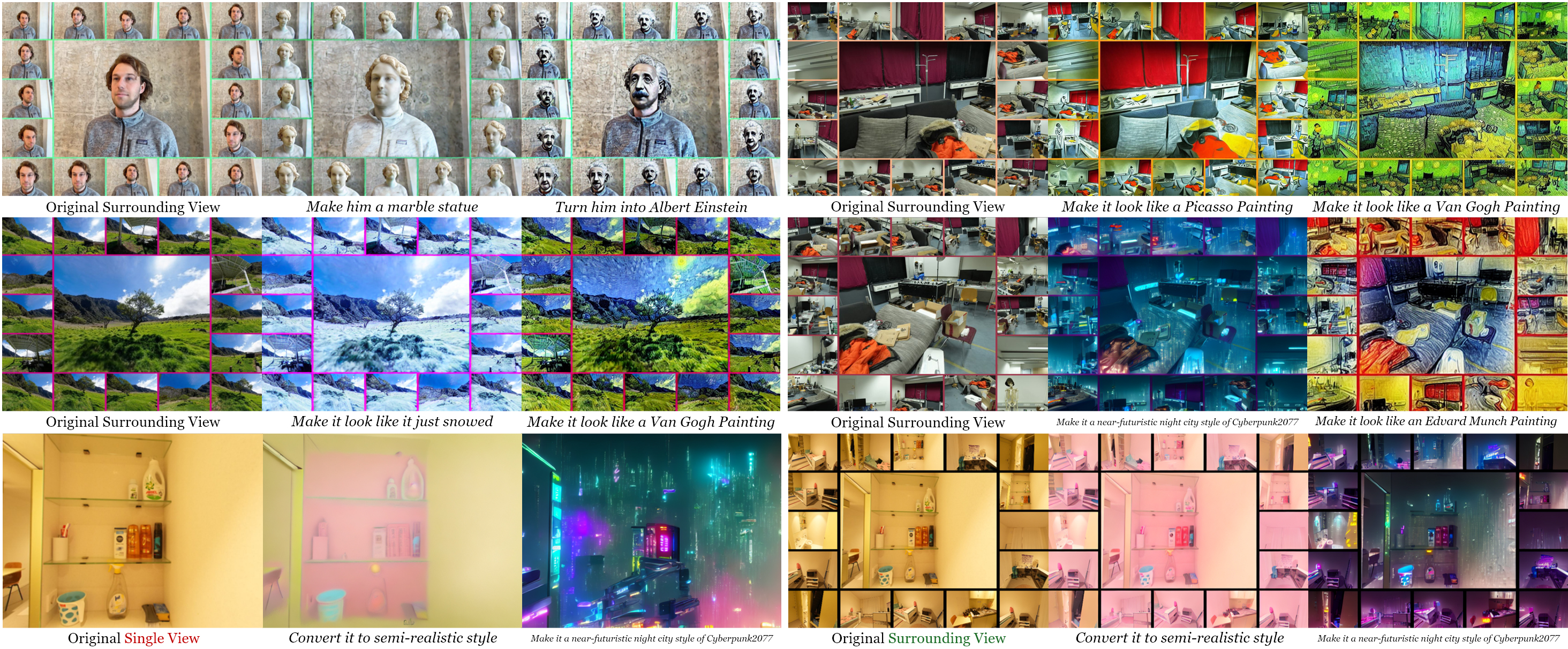}}

\caption{The pre-trained diffusion model \cite{ip2p} works as expected on surrounding views, by editing each sub-view in the instructed way individually but in a consistent style. Notably, as shown in the last row, the surrounding view enriches the context, making the diffusion model succeed in views that fail in single-view editing. }

\label{fig:evi-sur}
\end{figure*}
\begin{figure*}[t!]
\centering
\centerline{\includegraphics[width=1\linewidth]{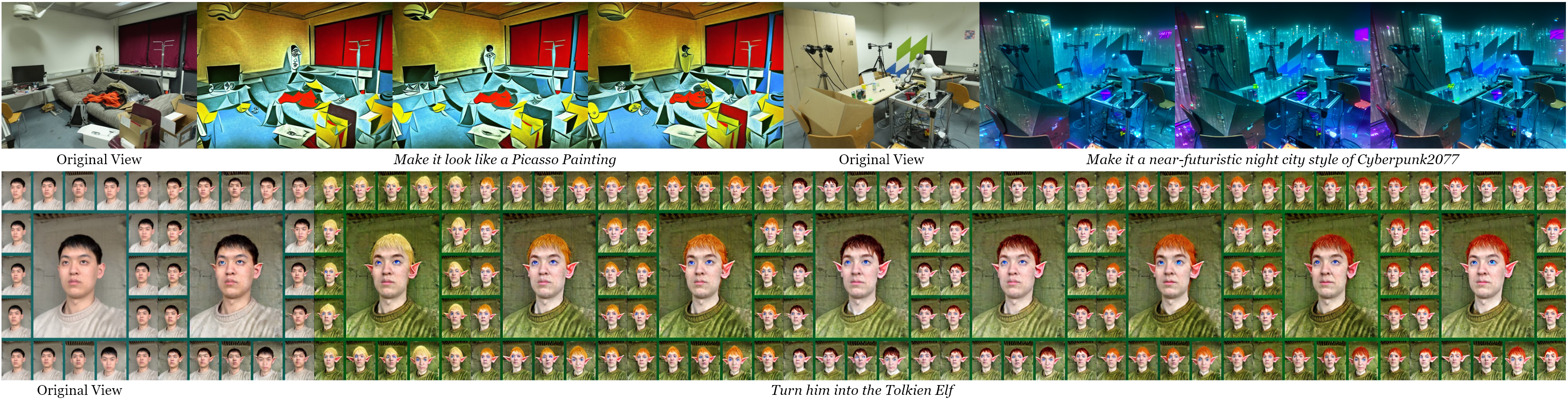}}

\caption{Even for the same view, generating from different noises does not necessarily lead to the consistent \ie, the same, edited result. Each column represents a generation from a noise different from other columns.}

\label{fig:evi-diffnoise}
\end{figure*}

\subsection{Diffusion Models Perform Well with Composed Images}
\label{sec:supp-evi-composed}
As shown in Fig. \ref{fig:evi-sur}, the pre-trained diffusion model \cite{ip2p}, though not directly trained in this pattern, still works as expected in surrounding views. It generates editing results for each sub-view individually while all of them also share a similar style, across various scenes, including indoor, outdoor, and face-forwarding scenes.

Notably, as shown in the last row, when editing a view with little context, directly editing the single view fails. Constructing a surrounding view using it as the main view, however, helps the diffusion model \cite{ip2p} to achieve successful editing. This shows the effects of surrounding views in achieving successful and consistent editing.
\subsection{Different Noises Lead to Varied Results }
As shown in Fig. \ref{fig:evi-diffnoise}, generation from different noises leads to completely different images, which is the fundamental constraint of all the baselines, which do not control the noise. Even with surrounding views, the diffusion model \cite{ip2p} still generates images in highly inconsistent ways. The diversity of the diffusion model under different noises is desirable in 2D generation and editing, but has to be controlled in 3D generation for consistency.

\section{Additional Ablation Study Analysis}

\subsection{`No Str. Noise' vs. `Only Sur. Views'}

Both variants do not have structured noise. Hence, the consistency-enforcing training in `No Str. Noise' forces the model to generate the same result from different noises, which leads to mode collapse and degrades the editing result towards blurred, averaged color. These negative effects of training in `No Str. Noise' leads to similar and even worse results and DFS than `Only Sur. Views' with no training.

\subsection{`Only Sur. Views' vs. `IN2N'} 
Tasks B,C,D are {\em style transfer}, specifically well supported by our current 2D diffusion model \cite{ip2p}. Our DFS metric, based on FID, uses a feature extractor with more tolerance for different style transfer results in the same image. Hence, even `IN2N' performs comparably with a slightly lower DFS.

By contrast, task A is a {\em general object-centric editing} with diversified editing manners -- different valid editing results can have jackets with completely different colors and styles. There can even be geometric changes in the clothing without surrounding views as context to constrain the editing, leading to a significantly worse DFS for `IN2N.'

\section{Discussion}
\subsection{Extension to Scene Generation}
The proposed \themodel primarily focuses on the distillation-guided 3D scene editing task. However, the core contributions -- structured noise, surrounding views, and consistency-enforcing training -- can also be extended to the scene generation task. For example, these components can be used in the refinement phase, when the shape of a scene is roughly determined. In this way, these components could help achieve consistent and high-fidelity generation, refining the shape with slight adjustments for more detailed and precise geometry. Compared with previous methods \cite{magic3d,prolificdreamer}, this method can generate scenes with detailed, high-fidelity textures and shapes, without mesh exportation or fixing geometry.

\subsection{Limitations}
This section discusses the limitations of \themodel, which are also the \emph{common challenges encountered by existing 3D scene editing methods}.

\paragraph{View-Dependent or Specular Effects.} Our \themodel pipeline performs consistency-enforcing training by warping and averaging between different views. This procedure enforces that each part of the scene ``looks the same'' in different views, \textit{i.e.,} is view-independent, making the edited scene unlikely to show view-dependent or specular effects. To preserve the ability to generate view-dependent effects, our \themodel has introduced a regularization loss that trades off between consistency and similarity to original \cite{ip2p} (detailed in Sec. \ref{sec:supp-impl-reg}). With this regularization, our \themodel could still achieve 3D consistency while allowing natural view-dependent effects. The baselines, though are not trained towards consistency or view-independence, only generate blurred results without notable effects, or even overfit to inconsistent editing with the view-dependency of NeRF.

\paragraph{Editing Capabilities Constrained by 2D Diffusion Models.} Our \themodel distills from the diffusion model \cite{ip2p} to edit scenes. Therefore, the editing ability, style, and diversity of \themodel are inherently constrained by \cite{ip2p}. Our \themodel edits a scene in a specific manner following \cite{ip2p}. For example, in the ``Vincent Van Gogh'' editing in Fig. \ref{fig:exp-suppl-dreameditor}, our \themodel, along with IN2N \cite{in2n} and ViCA \cite{vica} which use the same \cite{ip2p} for editing, shows a side effect that transfers the style of the image to Van Gogh's painting style. Moreover, we cannot support editing tasks on which the diffusion model cannot perform. Despite this common constraint among all the distillation-based methods, our \themodel successfully transfers most of the editing capabilities of the 2D diffusion model to 3D, by achieving high-quality and high-diversity 3D scene editing.

\paragraph{3D Understanding and Reasoning.} Though our \themodel is 3D-informed and 3D-aware with the additional input of 3D positional embedding -- already surpassing all the baselines -- it is unable to reason and understand the semantics of each part of 3D scenes. Therefore, while our \themodel can edit a view using the knowledge of the whole scene's shape (via 3D positional embedding) and appearance (through the surrounding view), it may still encounter multi-face issues or Janus problems. Specifically, it does not understand what the correct orientation of the face is, does not know that a person can only have one face, and thus cannot avoid this problem.

\paragraph{Shape Editing.} Some instructions for editing tasks may involve modifying the geometry or shape of a scene, \eg, ``give him a beard" creates a beard on the face. Like the baselines, our \themodel is designed to support simple shape editing tasks that can be achieved by slightly and gradually alternating the surface. For example, in the editing of ``give him a beard," our pipeline gradually ``grows" the beard's shape from the face's surface. Notice that both \themodel and the baselines cannot perform aggressive and complicated editing (\eg, removing an object while reconstructing the whole occluded part), or direction-related editing (\eg, performing ``lower down her arm'' for a scene of a person raising her arm requires a multi-view consensus on which direction the arm is moved to).

\paragraph{Efficiency.} In contrast to the diffusion training-free baselines such as~\cite{in2n,i3d23d,vica}, our \themodel needs additional training of a 2D diffusion model. This extends the editing duration, resulting in taking 12 hours to edit a scene in the IN2N dataset, and up to 24 hours to edit a large-scale indoor scene in the ScanNet++ dataset. However, as a trade-off against efficiency, our \themodel excels in achieving high-fidelity editing, surpassing all the training-free baselines.

\subsection{Future Directions}
\paragraph{Supporting Specular Effects.} One direction is to support specular effects and better view-dependency. This may need an improved formulation of consistency under specular reflections, or modeling the ambient environment.

\paragraph{3D Understanding for Scene Editing.} Another direction is to enable the diffusion model to understand and reason the semantics of a scene. Introducing a model that generates 3D semantic embeddings for each point in the scene allows for combining this information with the 3D positional embedding as the input to the diffusion model, potentially mitigating Janus problems.

\end{document}